# Flexible Auto-weighted Local-coordinate Concept Factorization: A Robust Framework for Unsupervised Clustering

Zhao Zhang, *Senior Member, IEEE,* Yan Zhang, Sheng Li, *Senior Member, IEEE,* Guangcan Liu, Dan Zeng, Shuicheng Yan, *Fellow, IEEE* and Meng Wang

**Abstract**— Concept Factorization (CF) and its variants may produce inaccurate representation and clustering results due to the sensitivity to noise, hard constraint on the reconstruction error and pre-obtained approximate similarities. To improve the representation ability, a novel unsupervised Robust Flexible Auto-weighted Local-coordinate Concept Factorization (RFA-LCF) framework is proposed for clustering high-dimensional data. Specifically, RFA-LCF integrates the robust flexible CF by clean data space recovery, robust sparse local-coordinate coding and adaptive weighting into a unified model. RFA-LCF improves the representations by enhancing the robustness of CF to noise and errors, providing a flexible constraint on the reconstruction error and optimizing the locality jointly. For robust learning, RFA-LCF clearly learns a sparse projection to recover the underlying clean data space, and then the flexible CF is performed in the projected feature space. RFA-LCF also uses a $L_{2,1}$-norm based flexible residue to encode the mismatch between the recovered data and its reconstruction, and uses the robust sparse local-coordinate coding to represent data using a few nearby basis concepts. For auto-weighting, RFA-LCF jointly preserves the manifold structures in the basis concept space and new coordinate space in an adaptive manner by minimizing the reconstruction errors on clean data, anchor points and coordinates. By updating the local-coordinate preserving data, basis concepts and new coordinates alternately, the representation abilities can be potentially improved. Extensive results on public databases show that RFA-LCF delivers the state-of-the-art clustering results compared with other related methods.

——————————— ◆ ———————————

## 1 INTRODUCTION

CLUSTERING high-dimensional data by effective representation is a fundamental issue in the areas of multivariate data analysis and data mining, etc. But the massive and ever-increasing real data usually have unfavorable features and various noise or errors that may decrease the representation results directly [38-42], so how to compute more descriptive and robust representations of original data is still challenging [29-31][57-60][62]. To address this issue, different representation learning methods can be used, among which matrix factorization is one widely-used method. In last decades, many effective factorization methods were proposed, of which *Vector quantization* (VQ) [3], *Principal Component Analysis* (PCA) [1], *Singular Value Decomposition* [2], *Nonnegative Matrix Factorization* (NMF) [4], and *Concept Factorization* (CF) [5] are classical methods for data analysis and representation. Among these factorization models, NMF and CF differ from others since the used nonnegative constraints on the factorization matrices can enable a non-subtractive combination of parts to form a whole, which can be regarded as a procedure of learning parts-based representations [4-5]. Note that in reality vital distinguishing features may be the key parts of faces (i.e., noses and eyes) in image data, topics in text data, or strokes in handwriting data, so the nonnegative constraint plays an essential role in data representation.

Given a nonnegative data matrix $X$, NMF and CF aim at decomposing it into the product of two or three nonnegative factors by minimizing the reconstruction error [4] [5]. One factor is the basis vectors capturing the higher-level features of data and each sample can be regarded as a linear combination of the bases approximately. The other factor contains the coefficients [4][5]. Then, the cluster label of each sample can be obtained from the linear coefficients. Due to the nonnegative constraint based additive reconstruction, NMF and its variants, e.g., *Projective NMF* (PNMF) [6], *Graph Regularized NMF* (GNMF) [7], *Constrained NMF* (CNMF) [8], *Graph Dual Regularization NMF* (DNMF) [41], *Parameter-less Auto-weighted Multiple Graph regularized NMF* (PAMGNMF) [50] and *Dual-graph Sparse NMF* (DSNMF) [49] are widely applied for characterizing and clustering the faces, documents and texts [45-47], etc. Although the enhanced results have been obtained, NMF and its variants still cannot handle data in the reproduc-

Z. Zhang is with School of Computer Science and Technology, Soochow University, Suzhou, China; also with School of Computer and Information, Hefei University of Technology, Hefei, China. (e-mail: cszzhang@gmail.com)
Y. Zhang is with School of Computer Science and Technology, Soochow University, Suzhou, China. (e-mail: zhangyan0712suda@gmail.com)
S. Li is with the Department of Computer Science, University of Georgia, 549 Boyd GSRC, Athens, GA 30602. (e-mail: sheng.li@uga.edu)
G. Liu is with the Nanjing University of Information Science and Technology, Nanjing, China. (e-mail: gcliu@nuist.edu.cn)
D. Zeng is with the Shanghai Institute of Advanced Communication and Data Science, Shanghai University, Shanghai 200444, China. (e-mail: dzeng@shu.edu.cn)
S. Yan is with the Department of Electrical and Computer Engineering, National University of Singapore, Singapore. (e-mail: eleyans@nus.edu.sg)
M. Wang is with the School of Computer and Information, Hefei University of Technology, Hefei, China. (e-mail: eric.mengwang@gmail.com)

ing kernel Hilbert space. To solve this issue, CF that aims at representing each sample using a linear combination of the cluster centers is derived. The major advantage of CF over NMF is that CF can be performed in any data representation space, thus it can be easily kernelized compared with the NMF based models. Note that CF can only reveal the global geometry of data space but cannot preserve the manifold structures. Toward handing this issue, several effective locality preserving CF based methods have been recently proposed, such as *Locally Consistent CF* (LCCF) [9], *Local Coordinate CF* (LCF) [10], *Graph-Regularized LCF* (GRLCF) [26], *Graph-regularized CF with Local Coordinate* (LGCF) [48] and *Dual-graph regularized CF* (GCF) [40]. To be specific, LCCF uses the graph Laplacian to smooth the representation and encode the geometrical information of the data space, which allows extracting the concepts with respect to the intrinsic manifold structures. Different from LCCF, LCF provides another effective method to preserve the locality by requiring the basis vectors to be as close to the original data as possible. Besides, it enables each data point to be represented by a linear combination with only a few nearby basis concepts so that the locality and sparsity are captured at the same time, i.e., local coordinate coding is incorporated with the locality preservation. As a combination of LCF and LCCF, both GRLCF and LGCF force the learned coefficients to be sparse and simultaneously keep the geometric structures of samples by integrating the local coordinate constraint and graph regularization, which improves the clustering result to some extent. GCF preserves the geometric sturctures of both data manifold and feature manifold simultaneously by using the dual-graph regularization strategy [40].

It is worth noting that existing CF methods still suffer from some drawbacks that potentially decrease the representation and clustering abilities. First, to keep the locality of the new representation, LCCF, GRLCF, LGCF and GCF need to search the neighbors of each data point by the $k$-neighborhood or $\varepsilon$-neighborhood, and pre-calculate the graph weights using a separate step before factorization. But estimating an optimal $k$ or $\varepsilon$ value is still a tricky issue in real applications [16-17], and fixing the same $k$ or $\varepsilon$ value for all samples is also unreasonable because real data usually have complex and different distributions [25]. The pre-calculated weights and graph Laplacian prior to the factorization process also cannot be ensured to be optimal for seeking the new representation of orginal data explicitly. Compared with LCCF and LCF, the recent LGCF and GRLCF incorporate the idea of local coordinate coding to capture the sparsity and locality simultaneously. But the aforementioned LCF, LCCF, GRLCF, LGCF and GCF still cannot encode the local geometrical structures in the basis concept space and new coordinate space jointly, especially in an adaptive manner. Second, the processes of searching neighbors, defining weights and performing factorization of aforementioned methods are performed in the original space, but real data usually have noise and unfavorable features that may cause negative effects on the representation results. Thus, it would be better to weight and represent data in a noise-removed clean space so that more accurate data representation can be obtained. Third, CF and its variants minimize the reconstruction error be-tween the original data $X$ and the product of three factors as a hard constraint for discovering the new representation. They assume that the new representation should lie in the nonnegative space and a linear combination of cluster centers should be able to represent each sample, but the hard constraint may be overfitted in real applications.

In this paper, we propose a novel robust locality preserving flexible factorization method to overcome existing shortcomings of aforementioned LCF, LCCF and GRLCF, and inherit their merits at the same time. The major contributions of this paper are summarized as follows:

(1) A novel and unsupervised framework called *Robust Flexible Auto-weighted Local-coordinate Concept Factorization* (RFA-LCF) is technically proposed. RFA-LCF aims to enhance the representation ability in threefold: (i) improving the robust properties of the factorization and coordinate coding to noise and corruption by subspace recovery; (ii) encoding the locality structures by adaptive weighting in the basis vectors space and new representation space; (iii) providing a more accurate flexible constraint on the reconstruction error. To integrate these innovations, we clearly incorporate the robust flexible CF, robust adaptive sparse local coordinate coding and auto-weighting into a unified model. The relationship analysis also illustrates that RFA-LCF is more general and powerful.

(2) For the robust flexible learning, RFA-LCF improves the representations in twofold. First, it enhances the robustness by seeking a sparse projection $P$ to obtain salient features of original data and remove noise from data by embedding jointly. Then, the factorization is performed in the projective feature space, which clearly differs from most existing methods that are usually performed in the original input space. We also use the sparse $L_{2,1}$-norm to encode the mismatch between the recovered data and its reconstruction. This is also different from most existing models with the Frobenius-norm that is usually sensitive to noise and outliers. While $L_{2,1}$-norm has been proven to be robust to noise and outliers and it can enforce the reconstruction error to be sparse in rows [11][51][52], which has the potential to minimize the factorization error. Second, RFA-LCF introduces a soft and flexible penalty term on reconstruction error by relaxing the existing assumption that each sample can be represented by a linear combination of cluster centers. This operation can avoid the possible overfitting issue and handle the data sampled from a nonlinear manifold potentially [43][44].

(3) To encode the locality and sparsity more accurately, we integrate the adaptive weighting with the robust flexible CF by discovering not only the manifold structures in the projective feature space, but also the localities of the basis concepts and new representations in an adaptive manner. By sharing the adaptive weights in the projective feature space, basis vector space and new representation space, the encoded similarities and locality can be potentially more reliable. The auto-weighting can also avoid the tricky issue of specifying a fixed neighborhood size or ball radius. Based on the adaptive weights and projective features, RFA-LCF can also perform the robust adaptive locality preserving sparse local coordinate coding to represent data by using a few most nearby basis concepts.

This paper is outlined as follows. Section 2 briefly reviews the related work. Section 3 presents the problem, convergence analysis and complexity of our RFA-LCF. In Section 4, we show the connections between our method and other models. Section 5 shows the settings and results. Finally, the paper is concluded in Section 6.

## 2 RELATED WORK

We briefly review the related CF, LCCF and LCF methods.

### 2.1 Matrix factorization and CF

CF is a classical unsupervised factorization method for data representation. Given $X = [x_1, x_2, ..., x_N] \in \mathbb{R}^{d \times N}$, where $x_i, i \in 1, 2, ..., N$ are sample vectors, $N$ denotes the number of samples and $d$ the original dimension of each sample. Denoting $U \in \mathbb{R}^{d \times r}$ and $V^T \in \mathbb{R}^{r \times N}$ as two nonnegative matrices, $UV^T \in \mathbb{R}^{d \times N}$ is the approximation to the data $X$, where the rank $r$ of the factorization is a constant. By representing each basis using a nonnegative linear combination of $x_i$, i.e., $\sum_{i=1}^{N} w_{ij} x_i$, where $w_{ij} \geq 0$, then CF aims at calculating the approximate relation as $X \approx XWV^T$. Thus, CF minimizes the following objective function:

$$\min_{W, V} \left\| X - XWV^T \right\|^2, \; s.t. \, W, V \geq 0, \tag{1}$$

where $W = [w_{ij}] \in \mathbb{R}^{N \times r}$, $XW$ approximates the bases, $V^T$ is the new representation, and $V^T$ is the transpose of $V$.

**Theorem 1.** For $X, W, V \geq 0$, the objective function of CF is nonincreasing under the following updating rules [5]:

$$w_{jk}^{t+1} \leftarrow w_{jk}^t \frac{(KV)_{ik}}{(KWV^T V)_{ik}}, \; v_{jk}^{t+1} \leftarrow v_{jk}^t \frac{(KW)_{ik}}{(VW^T KW)_{ik}}, \tag{2}$$

where $K = X^T X$. After the convergence of CF, the new representation $V^T$ of the original data can be obtained.

### 2.2 Locally Consistent CF (LCCF)

Differernt from the regular CF, LCCF learns the manifold preserving representations of the original data by adding a geometrically based regularizer. LCCF first constructs a graph $G(R, E)$ with $N$ nodes over $X$, where each vertex in the vertex set $R$ corresponds to $x_i$, and the edge weight $S_{ij}$ connecting $x_i$ and $x_j$ can be defined as

$$S_{ij} = \begin{cases} \dfrac{x_i^T x_j}{\|x_i\| \|x_j\|}, & \text{if } x_i \in N_k(x_j) \text{ or } x_j \in N_k(x_i) \\ 0, & \text{otherwise} \end{cases}, \tag{3}$$

where $N_k(x_i)$ is the set including the $k$ nearest neighbors of $x_i$. The regularization term $\Re$ can then be defined as

$$\Re = 1/2 \sum_{i,j=1}^{N} \left\| v_i - v_j \right\|^2 S_{ij} = tr(V^T L V), \tag{4}$$

where $v_i$ is the $i$-th column of $V$, graph Laplacian $L = D - S$, $D$ is a diagonal matrix whose entries are column sums of $S$, i.e., $D_{ii} = \sum_j S_{ij}$. By representing each basis by a nonnegative linear combination of $x_i$, i.e., $\sum_{i=1}^{N} w_{ij} x_i$, where $w_{ij} \geq 0$, LCCF solves the following objective function:

$$\min_{W, V} \left\| X - XWV^T \right\|^2 + \lambda tr(V^T L V), \; s.t. \, W, V \geq 0, \tag{5}$$

where $\lambda \geq 0$ denotes a regularized weighting factor.

**Theorem 2.** For $X, W, V \geq 0$, the objective function of LCCF in Eq.(5) is nonincreasing under the following multiplicative updating rules [9]:

$$w_{jk}^{t+1} \leftarrow w_{jk}^t \frac{(KV)_{ik}}{(KWV^T V)_{ik}}, \; v_{jk}^{t+1} \leftarrow v_{jk}^t \frac{(KW + \lambda SV)_{ik}}{(VW^T KW + \lambda DV)_{ik}}. \tag{6}$$

### 2.3 Local Coordinate CF (LCF)

LCF is motivated by the idea of local coordinate coding. More specifically, LCF takes the locality constraints into account, considers the anchor points $u_j = \sum_i w_{ij} x_i$ and the coordinates for each sample over each column of $V$ with respect to the anchor points. Then, LCF defines the following constraint term to measure the locality and sparsity penalties between the anchor point $u_r$ and $x_i$:

$$\sum_{r=1}^{R} |v_{ri}| \|u_r - x_i\|^2 = \sum_{r=1}^{R} |v_{ri}| \left\| \sum_{j=1}^{N} w_{jr} x_j - x_i \right\|^2, \tag{7}$$

Thus, LCF minimizes the following objective function:

$$\min_{W, V} \left\| X - XWV \right\|^2 + \lambda \sum_{i=1}^{N} \sum_{r=1}^{R} |v_{ri}| \left\| \sum_{j=1}^{N} w_{jr} x_j - x_i \right\|^2, \tag{8}$$

where $\lambda \geq 0$ is a weighting parameter. By Eq. (8), LCF tries to represent $x_i$ by using only a few nearby anchor points so that the sparsity and local structure can be preserved.

**Theorem 3.** For $X, W, V \geq 0$, the objective function of LCF in Eq.(8) is nonincreasing under the following multiplicative updating rules [10]:

$$w_{jk}^{t+1} \leftarrow w_{jk}^t \frac{\left( KV^T + \lambda \sum_{i=1}^{N} X^T x_i e^T D_i \right)_{ik}}{\left( KWVV^T + \lambda \sum_{i=1}^{N} KWD_i \right)_{ik}},$$

$$v_{jk}^{t+1} \leftarrow v_{jk}^t \frac{2(\lambda+1)(W^T K)_{ik}}{(2W^T KWV + \lambda A + \lambda B)_{ik}} \tag{9}$$

where $e$ is a column vector with all ones, $A$ is a $R \times N$ matrix whose rows are $a^T = (diag(K))^T$, and $B$ is a $R \times N$ matrix whose columns are $b = diag(W^T KW)$.

## 3 ROBUST FLEXIBLE AUTO-WEIGHTED LOCAL-CORDINATE CONCEPT FACTORIZATION

### 3.1 The Objective Function

We introduce the formulation of RFA-LCF in this section. The main idea of RFA-LCF is to improve the data representation ability by improving the robustness properties to noise and errors by subspace recovery, enhancing the similarities by robust sparse local-coordinate coding and adaptive weighting learning, and providing a more flexible reconstructive factorization error. Given a data matrix $X = [x_1, x_2, ..., x_N] \in \mathbb{R}^{d \times N}$, RFA-LCF jointly learns a $L_{2,1}$-norm based sparse subspace projection $P \in \mathbb{R}^{d \times d}$ to remove the noise and outliers in data by embedding $X$ onto it directly. Then, it factorizes the recovered clean data $P^T X$ to obtain the nonnegative matrices $W \in \mathbb{R}^{N \times R}$ and $V^T \in \mathbb{R}^{R \times N}$, and uses the product $XWV^T \in \mathbb{R}^{d \times N}$ to approximate the clean data $P^T X$ for more descriptive and robust representations. For robust flexible CF, RFA-LCF clearly sets the factorization based on the recovered clean data as $P^T X = \hbar(X, W, V) + P_0^T X$, where $\hbar(X, W, V)$ is a transform function for factorizing data. Assume that $\hbar(X, W, V)$ is the linear regression function $XWV^T$, the residue $P_0^T X$ will be able to encode the mismatch between $P^T X$ and $XWV^T$ in a soft manner by relaxing the hard constraint $P^T X = XWV^T$, and can also avoid the possible overfitting problem in the reconstruction. It is worth noting that virtually all existing CF based algorithms directly minimize the reconstruction error between the original data and the product, and therefore they may suffer from the potential overfitting issue in reality. To make the residue easily understandable, we introduce a bias term into it, i.e., $eb^T$, where $e$ is a column vector of all ones and $b \in \mathbb{R}^{d \times 1}$ is a bias vector that also corresponds to the matching error. Since the sparse $L_{2,1}$-norm is proved to be more suitable to encode the reconstruction error and more robust to noise than traditional Frobenius-norm [11][51][52], we impose $L_{2,1}$-norm to encode the residue, i.e., $\left\| X^T P + eb^T - VW^T X^T \right\|_{2,1}$. To encode the neighborhood and similarities more reliably, RFA-LCF defines the manifold structures jointly over the clean data $P^T X$, basis concepts $XW$ and coordinates $V^T$ in an adaptive manner by minimizing the joint reconstruction errors explicitly, i.e., $\left\| P^T X - P^T XQ \right\|_F^2 + \left\| W^T - W^T Q \right\|_F^2 + \left\| V^T - V^T Q \right\|_F^2$, where $Q$ is the adaptive reconstruction weight matrix to be calculated. RFA-LCF also involves the robust adaptive locality preserving coordinate coding to represent the data by using a few most nearby basis concepts, which can potentially enable the factorization process to have enhanced representation and clustering abilities. These discussions lead to the following objective function for RFA-LCF:

$$O = \min_{W, V, Q, P, b} \left\| X^T P + eb^T - VW^T X^T \right\|_{2,1}$$
$$+ \alpha f(W, V) + \beta g(Q) + \gamma \left\| P \right\|_{2,1} . \qquad (10)$$
$$s.t. \ W, V, Q \geq 0, \ Q_{ii} = 0, i = 1, 2, ..., N$$

where $W, V, Q \geq 0$ are the nonnegative constraints, $Q_{ii} = 0$ is added to avoid the trivial solution $Q = I$, and $\alpha, \beta, \gamma > 0$ are trade-off parameters. For the effects of $L_{2,1}$-norm [11] [51][52], minimizing the $L_{2,1}$-norm based flexible residue will have the potential to reduce the reconstruction error. $f(W, V)$ is the robust adaptive neighborhood preserving locality and sparsity constraint term and $g(Q)$ is the auto-weighted learning term, which are defined as

$$f(W, V) = \sum_{i=1}^{N} \sum_{r=1}^{R} \left| v_{ri}^T \right| \left\| \sum_{j=1}^{N} w_{jr} P^T x_j - P^T x_i \right\|^2$$
$$g(Q) = \left\| \begin{pmatrix} P^T X \\ W^T \\ V^T \end{pmatrix} - \begin{pmatrix} P^T X \\ W^T \\ V^T \end{pmatrix} Q \right\|_F^2 . \qquad (11)$$

The term $f(W, V)$ correlates the anchor points and the coordinates in coordinate coding, and the auto-weighted learning term $g(Q)$ keeps the reconstruction relationship and manifold structures of clean data $P^T X$, basis vectors $W^T$ and new representation $V^T$ jointly. To highlight and show the benefits of involving the two terms $f(W, V)$ and $g(Q)$, next we briefly discuss the sum of them:

$$\alpha f(W, V) + \beta g(Q) = \alpha \sum_{i=1}^{N} \sum_{r=1}^{R} \left| v_{ri}^T \right| \left\| \sum_{j=1}^{N} w_{jr} P^T x_j - P^T x_i \right\|^2$$
$$+ \left\| \begin{pmatrix} \sqrt{\beta} P^T X \\ \sqrt{\beta} W^T \\ \sqrt{\beta} V^T \end{pmatrix} - \begin{pmatrix} \sqrt{\beta} P^T X \\ \sqrt{\beta} W^T \\ \sqrt{\beta} V^T \end{pmatrix} Q \right\|_F^2 , \qquad (12)$$

from which one can find that the above coordinate coding process in our formulation is clearly different from that of LCF in twofold. First, the neighborhood relationship is encoded explicitly in an adaptive manner by integrating the reconstruction error $\left\| \sqrt{\beta} W^T - \sqrt{\beta} W^T Q \right\|_F^2 + \left\| \sqrt{\beta} V^T - \sqrt{\beta} V^T Q \right\|_F^2$ based on the basis concept vectors $W^T$ and new coordinates $V^T$ into the local coordinate coding in our problem, which may produce more accurate coordinates and representations than the LCF potentially. Second, our RFA-LCF performs the robust adaptive sparse local coordinate coding in the recovered clean data space spanned by using $P$ to represent data, while LCF performs the coordinate coding in the original data space that usually contains noise. Thus, RFA-LCF involves a robust adaptive neighborhood preserving locality and sparsity constrained penalty between the anchor point $u_r$ and $x_i$, compared with LCF.

Note that our RFA-LCF can be performed alternately among the following three steps. By updating the projective data, adaptive weights, basis concepts and new coordinates alternately in an adaptive manner, the representation ability can be potentially improved.

### (1) Robust Flexible Auto-weighted Local-coordinate CF:

When the projection $P$ and adaptive weight matrix $Q$ are fixed, we can focus on the robust flexible adaptive local-coordinate CF for representing data. With $P$ and $Q$ fixed, we have the following reduced formulation:

$$\min_{W, V} \left\| X^T P + eb^T - VW^T X^T \right\|_{2,1} + \alpha \sum_{i=1}^{N} \sum_{r=1}^{R} \left| v_{ri}^T \right| \left\| \sum_{j=1}^{N} w_{jr} P^T x_j - P^T x_i \right\|^2$$
$$+ \beta \left\| \begin{pmatrix} W^T \\ V^T \end{pmatrix} - \begin{pmatrix} W^T \\ V^T \end{pmatrix} Q \right\|_F^2 , \ s.t. \ W, V \geq 0 \qquad (13)$$

where $\alpha$ and $\beta$ trade off the robust adaptive sparse local-coordinate coding term and adaptive locality preserving term. Due to the adaptive weights $Q$, the neighborhoods within the basis concept vectors $XW$ and new coordinates $V^T$ can be preserved in an adaptive manner clearly. After $W$ and $V$ are obtained, we can update the neighborhood preserving sparse projection $P$ for subspace recovery.

### (2) Robust Subspace Recovery by Sparse Projection:

We focus on computing the robust $L_{2,1}$-norm regularized projection for recovering the subspace and removing the noise from data in this step, with $W$, $V$ and $Q$ known. The sub-problem involved can be formulated as

$$\min_{P} \left\| X^T P + eb^T - VW^T X^T \right\|_{2,1} + \beta \left\| P^T X - P^T XQ \right\|_F^2 + \gamma \left\| P \right\|_{2,1} , \qquad (14)$$

where the minimization of $\left\| P^T X - P^T XQ \right\|_F^2$ can preserve the neighborhood information of embedded data $P^T X$ in the projective feature space clearly. After the recovered subspace is updated by $P$, we can return it for the robust flexible adaptive local-coordinate CF and re-weighting.

*(3) Auto-weighted Reconstruction Graph Learning:*

When the sparse projection $P$ and factors $W$, $V$ are known, we can perform the auto-weighting learning by sharing the weight matrix $Q$ in projective feature space, basis vectors space and representation space, which can preserve the geometrical structures of embedded data, basis vectors and new representations explicitly and adaptively. The auto-weighted learning process is formulated as

$$\min_{Q} \left\| \begin{pmatrix} \sqrt{\beta} P^T X \\ \sqrt{\beta} W^T \\ \sqrt{\beta} V^T \end{pmatrix} - \begin{pmatrix} \sqrt{\beta} P^T X \\ \sqrt{\beta} W^T \\ \sqrt{\beta} V^T \end{pmatrix} Q \right\|_F^2, \ s.t. \ Q \ge 0, \ Q_{ii} = 0, \quad (15)$$

from which the adaptive weights in $Q$ can be obtained, where the entry $Q_{ij}$ measures the contribution of $x_j$ to reconstruct each $x_i$. That is, the larger $Q_{i,j}$ is, the closer between $x_j$ and $x_i$ in terms of similarity or distance is. By updating the adaptive weights, robust projection and factorization matrices alternately, RFA-LCF can ensure the learnt weights to be optimal for the data representations. Note that an early version of this work was presented in [57]. This paper has further provided the detailed analysis of the formulation, convergence analysis, computational complexity analysis and relationship analysis, and moreover conducts a thorough experimental evaluation on the tasks of data representation and clustering.

## 3.2 Optimization

We describe how to optimize the proposed objective function in Eq.(10). As there are several variables in the problem and the involved variables depend on each other, the objective function of RFA-LCF cannot be solved directly. In this paper, we follow the common procedures to update the variables alternately. Let $O$ be the objective function of our RFA-LCF, $R = X^T P + e b^T - V W^T X^T = [r^1, r^2, ..., r^N]$ and $M \in \mathbb{R}^{N \times N}$ be a diagonal matrix with the entries being $m_{ii} = 1 / \left( 2 \|r^i\|_2 \right)$, $i = 1, 2, ..., N$. Based on the properties of the L$_{2,1}$-norm [11][51-52], we can have $\left\| X^T P + e b^T - V W^T X^T \right\|_{2,1} = 2 tr \left( \left( P^T X + b e^T - X W V^T \right) M \left( X^T P + e b^T - V W^T X^T \right) \right)$. By taking the derivative of the problem in Eq.(10) $w.r.t.$ the bias $b$, and setting the derivative to zero, we can easily obtain

$$\frac{\partial O}{\partial b} = P^T X M e + b e^T M e - X W V^T M e = 0$$
$$\Rightarrow b = \left( X W V^T M e - P^T X M e \right) \big/ N_M^+, \quad (16)$$

where $N_M^+ = e^T M e = \sum_i m_{ii}$ is a constant. Thus, the flexible residue $\wp_{loss}$ can be rewritten as

$$\wp_{loss} = X^T P + e b^T - V W^T X^T$$
$$= X^T P + e \left( e^T M V W^T X^T - e^T M X^T P \right) \big/ N_M^+ - V W^T X^T$$
$$= H_e X^T P - \left( V W^T X^T - e e_M^T V W^T X^T / N_M^+ \right), \quad (17)$$
$$= H_e X^T P - H_e V W^T X^T$$

where $H_e = I - e e_M^T / N_M^+$ and $e_M^T = e^T M = [m_{11}, m_{22}, ..., m_{NN}] \in \mathbb{R}^N$. Then, we can have $\|P\|_{2,1} = tr \left( 2 P^T S P \right)$, $\left\| H_e X^T P - H_e V W^T X^T \right\|_{2,1} = tr \left( 2 \left( P^T X - X W V^T \right) H_e^T M H_e \left( X^T P - V W^T X^T \right) \right)$, where $S$ is a diagonal matrix with the entries being $s_{ii} = 0.5 / \|P^i\|_2$ and $P^i$ is the $i$-th row vector of $P$. Suppose each $\left( H_e X^T P - H_e V W^T X^T \right)^i \ne 0$ and $P^i \ne 0$ over each index $i$, let $H = \left( X^T P, W, V \right)^i$ and by

substituting Eq.(17) into Eq.(10), we have the following matrix trace based optimization problem:

$$\min_{W, V, Q, P, M, S} tr \left( \left( P^T X - X W V^T \right) H_e^T M H_e \left( X^T P - V W^T X^T \right) \right)$$
$$+ \alpha tr \left( \sum_{i=1}^N \left( P^T x_i e^T - P^T X W \right) L_i \left( P^T x_i e^T - P^T X W \right)^T \right), \quad (18)$$
$$+ \beta tr \left( H G H^T \right) + \gamma tr \left( P^T S P \right)$$
$$s.t. \ W, V, Q \ge 0, \ Q_{ii} = 0$$

where $L_i = diag \left( |v_i| \right)$ and $G = (I - Q)(I - Q)^T$. Then, the optimization of our RFA-LCF can be described as follows:

**(1) Fix $P$ and $Q$, update the nonnegative factors $W$, $V$:**
We first show how to optimize $W$ and $V$. Let $\psi_{jr}$ and $\phi_{jr}$ be the Lagrange multipliers for nonnegative constraints $W \ge 0$ and $V \ge 0$ respectively, $\Psi = \left[ \psi_{jr} \right]$ and $\Phi = \left[ \phi_{jr} \right]$, the Lagrange function $\mathcal{L}_1$ of Eq.(18) can be constructed as

$$\mathcal{L}_1 = tr \left( \left( P^T X - X W V^T \right) H_e^T M H_e \left( X^T P - V W^T X^T \right) \right)$$
$$+ \alpha tr \left( \sum_{i=1}^N \left( P^T x_i e^T - P^T X W \right) L_i \left( P^T x_i e^T - P^T X W \right)^T \right). \quad (19)$$
$$+ \beta tr \left( H G H^T \right) + tr \left( \Psi W^T \right) + tr \left( \Phi V \right)$$

By taking the partial derivatives of $\mathcal{L}_1$ with respect to $W$ and $V$ respectively, we can obtain

$$\partial \mathcal{L}_1 / \partial W = \left( 2 X^T X W V^T H_e^T M H_e V - 2 X^T P^T X H_e^T M H_e V \right)$$
$$+ \alpha \left( \sum_{i=1}^N -2 X^T P P^T x_i e^T L_i + 2 X^T P P^T X W L_i \right), \quad (20)$$
$$+ \beta \left( 2 G W \right) + \Psi$$

$$\partial \mathcal{L}_1 / \partial V = \left( 2 H_e^T M H_e V W^T X^T X W - 2 H_e^T M H_e X^T P X W \right)$$
$$+ \alpha \left( A - 2 W^T X^T P P^T X + B \right) + \beta \left( 2 G V \right) + \Phi. \quad (21)$$

By using the Karush-Kuhn-Tucker conditions [53-54], i.e., $\psi_{jr} w_{jr} = 0$ and $\phi_{jr} v_{jr} = 0$, and denoting $C = H_e^T M H_e$, we can easily obtain the following two equations:

$$\left( K W V^T C V - X^T P^T X C V \right) w_{jr}$$
$$+ \alpha \left( \sum_{i=1}^N -X^T P P^T x_i e^T L_i + X^T P P^T X W L_i \right) w_{jr} + \beta \left( G W \right) w_{jr} = 0, \quad (22)$$

$$\left( 2 C V W^T K W - 2 C X^T P X W \right) v_{jr} + \alpha \left( A - 2 W^T X^T P P^T X + B \right) v_{jr}$$
$$+ \beta \left( 2 G V \right) v_{jr} = 0 \quad (23)$$

where $A$ represents a $R \times N$ matrix whose rows are $a^T = \left( diag \left( X^T P P^T X \right) \right)$, and $B$ is a $R \times N$ matrix whose columns are $\upsilon = diag \left( W^T X^T P P^T X W \right)$. Note that the above equations can lead to following multiplicative updating rules for the basis vectors $W$ and new representation $V$:

$$w_{jr} \leftarrow w_{jr} \frac{\left( X^T P^T X C V + \alpha \sum_{i=1}^N X^T P P^T x_i e^T L_i \right)_{jr}}{\left( K W V^T C V + \alpha \sum_{i=1}^N X^T P P^T X W L_i + \beta G W \right)_{jr}}, \quad (24)$$

$$v_{jr} \leftarrow v_{jr} \frac{\left( 2 C X^T P X W + 2 \alpha K W^T \right)_{jr}}{\left( 2 C V W^T K W + \alpha \left( A^T + B^T \right) + 2 \beta G V \right)_{jr}}. \quad (25)$$

**(2) Fix $W$, $V$, and $Q$, update the sparse projection $P$:**
In this step, we show how to optimize $P$. By removing the terms that are irrelevant to variable $P$ from Eq.(18), we can obtain the following reduced formulation:

$$\min_P J(P) = tr\left(\left(P^T X - X W V^T\right) H_e^T M H_e \left(X^T P - V W^T X^T\right)\right)$$
$$+ \alpha tr\left(\sum_{i=1}^N \left(P^T x_i e^T - P^T X W\right) L_i \left(P^T x_i e^T - P^T X W\right)^T\right) \quad . (26)$$
$$+ \beta tr\left(P^T X G X^T P\right) + \gamma tr\left(P^T S P\right)$$

By taking the derivative of $J(P)$ w.r.t. $P$ and setting the derivative $\partial J(P)/\partial P$ to zero, we have

$$\partial J(P)/\partial P = X C X^T P - X V W^T X^T + \beta X G X^T P + \gamma S P$$
$$+ \alpha \sum_{i=1}^N \left(x_i e^T - X W\right) L_i \left(x_i e^T - X W\right)^T P = 0. \quad (27)$$

Thus, we can easily update the projection matrix $P$ as

$$P = \left(X\left(C + \alpha \Xi + \beta G\right) X^T + \gamma S\right)^{-1} X C V W^T X^T, \quad (28)$$

where $\Xi = (E - W) L (E - W)^T$ and $E$ is an $N \times R$ matrix with all ones. After $P$ is updated in each iteration, we can use it together with $W$ and $V$ to compute the weight matrix $Q$.

**(3) Fix $W$, $V$, and $P$, update the adaptive weights $Q$:**
By removing the irrelevant terms to $Q$ from Eq.(18), we can obtain the following reduced formulation:

$$\min_Q J(Q) = \beta tr\left(H\left(I - Q\right)\left(I - Q\right)^T H^T\right), s.t. Q \geq 0, Q_{ii} = 0, \quad (29)$$

where $H = \left(X^T P, W, V\right)^T$. Let $\tau_{ij}$ be the Lagrange multiplier for the nonnegative constraint $Q \geq 0$, and $\Gamma = \left[\tau_{ij}\right]$, the Lagrange function $\mathcal{L}_2$ of Eq.(29) can be constructed as

$$\mathcal{L}_2 = \beta tr\left(H\left(I - Q\right)\left(I - Q\right)^T H^T\right) + tr\left(\Gamma Q^T\right). \quad (30)$$

By taking the derivative of $\mathcal{L}_2$ with respect to $Q$, and using the KKT condition $\tau_{ii} q_{ii} = 0$, we can easily obtain

$$\partial \mathcal{L}_2 / \partial Q = 2\beta\left(H^T H Q - H^T H\right) + \Gamma$$
$$2\beta\left(H^T H Q - H^T H\right) q_{ij} = 0 \quad , \quad (31)$$

which leads to the following updating rule for $Q$:

$$q_{ij} \leftarrow q_{ij} \frac{\left(H^T H\right)_{ij}}{\left(H^T H Q\right)_{ij}} \text{ and } q_{ii} = 0. \quad (32)$$

After the adaptive weight matrix $Q$ is obtained in each iteration, we can return it to further update $W$, $V$ and $P$.

To present our RFA-LCF completely, we summarize its optimization procedures in Algorithm 1, where the diagonal matrices $M$ and $S$ are initialized to be the identity matrices similarly as [11] so that each vector $P^i \neq 0$ and $\left(H_e X^T P - H_e V W^T X^T\right)^i \neq 0$ over index $i$ can be satisfied in the iterative optimizations. The convergence condition is set to $\left\| V^{t+1} - V^t \right\|_F \leq \varepsilon$, where $\varepsilon$ is a small number set to 0.001 in this paper. It can measure the divergence between two sequential representations and ensure the representation result will not change drastically, since the representation $V$ is the major variable computed for data clustering.

### 3.3 Convergence Analysis
We show the convergence analysis of our RFA-LCF. We present Theorem 4 regarding the above iterative updating rules, which ensures the convergence of the iterations and the final solution will be the local optimum.

**Theorem 4:** The problem of RFA-LCF in Eq.(18) is non-increasing under the presented updating rules.

---

| **Algorithm 1: Our proposed RFA-LCF framework** |
|---|
| **Inputs:** Original data matrix $X$, constant $r$ (rank of the factorization), and the tuning parameters $\alpha, \beta, \gamma$. |
| **Initialization:** Initialize the weight matrix $Q$ by the cosine similarity; Initialize $W$ and $V$ as random matrices; Initialize $M$, $S$ and $P$ to be the identity matrices; $t = 0$. |
| ***While not converged do*** |
| 1. Update $W$ and $V$ by Eqs.(24) and (25); |
| 2. Update the robust projection $P$ by Eq. (28) ; |
| 3. Update the adaptive weight matrix $Q$ by Eq. (32) ; |
| 4. Update the diagonal matrices $M$ and $S$ accordingly; |
| 5. Check for convergence: suppose that $\left\| V^{t+1} - V^t \right\|_F \leq \varepsilon$, stop; else $t = t + 1$. ***End while*** |
| **Output:** New representation $\left(V^T\right)^* = \left(V^T\right)^{t+1}$, adaptive weight matrix $Q^* = Q^{t+1}$ and robust projection $P^* = P^{t+1}$. |

---

To prove the Theorem 4, we use a similar convergence proof method as [4][10] by adopting auxiliary functions to assist the analysis. We show the definition of the auxiliary function and its property as follows.

**Definition 1:** $\Delta(x, x')$ is an auxiliary function for $F(x)$ if the following conditions are satisfied:

$$\Delta(x, x') \geq F(x), \Delta(x, x) = F(x). \quad (33)$$

**Lemma 1:** If $\Delta$ denotes an auxiliary function, then $F$ is non-increasing under the update:

$$x^{t+1} = \arg\min_x \Delta(x, x'). \quad (34)$$

**Proof:** $F\left(x^{t+1}\right) \leq \Delta\left(x^{t+1}, x'\right) \leq \Delta\left(x', x'\right) = F\left(x'\right)$.

Note that the equality $F\left(x^{t+1}\right) = F\left(x'\right)$ holds only if $x'$ is a local minimum of $\Delta(x, x')$. By iterating the updates, we can obtain a sequence of estimates that converge to a local minimum $x_{\min} = \arg\min_x F(x)$. Next, we define an auxiliary function for our problem and use Lemma 1 to show that the minimum of objective function is exactly our update rule, and thus the Theorem 4 can be proved.

We prove the convergence of the updating rule in Eq.(24) firstly. For any entry $w_{ij}$ in $W$, let $F_{w_{ij}}$ be the part of objective function relevant to $w_{ij}$, $F_{w_{ij}}$ can be defined as

$$\min_W tr\left(\left(P^T X - X W V^T\right) H_e^T M H_e \left(X^T P - V W^T X^T\right)\right)$$
$$+ \alpha tr\left(\sum_{i=1}^N \left(P^T x_i e^T - P^T X W\right) L_i \left(P^T x_i e^T - P^T X W\right)^T\right). (35)$$
$$+ \beta tr\left(W G W^T\right), \ s.t. \ W \geq 0$$

Since the update is essentially element-wise, it is sufficient to show each $F_{w_{ij}}$ is non-increasing under the update rule of Eq.(24). To prove it, we can define the following auxiliary function $\Delta$ for $F_{w_{ij}}$.

**Lemma 2:** The following function is an auxiliary function for $F_{w_{ij}}$, which is only relevant to variable $w_{ij}$:

$$\Delta\left(w, w_{ij}^t\right) = F_{w_{ij}}\left(w_{ij}^t\right) + F'_{w_{ij}}\left(w_{ij}^t\right)\left(w - w_{ij}^t\right) +$$
$$\frac{\left(\left(K W V^T C V + \alpha \sum_{i=1}^N X^T P P^T X W L_i + \beta G W\right)_{ij}\right)}{w_{ij}^t}\left(w - w_{ij}^t\right)^2. (36)$$

**Proof:** The Taylor series expansion of $F_{w_{ij}}$ is described as $F_{w_{ij}}(w) = F_{w_{ij}}\left(w_{ij}^t\right) + F'_{w_{ij}}\left(w_{ij}^t\right)\left(w - w_{ij}^t\right) + 1/2 F''_{w_{ij}}\left(w - w_{ij}^t\right)^2$. Since

$$\left(KWV^T CV + \alpha \sum_{i=1}^{N} KWL_i\right)_{ij} + \beta GW_{ij}$$

$$= \sum_k (KW)_{ik}\left(V^T CV\right)_{kj} + 2\alpha \sum_{i=1}^{N}\sum_k \left(X^T PP^T XW\right)_{ik}(L_i)_{kj}$$

$$+ \beta G_{ii}W_{ij} \geq (KW)_{ij}\left(V^T CV\right)_{jj} + 2\alpha \sum_{i=1}^{N}\left(X^T PP^T XW\right)_{ij}(L_i)_{jj}$$

$$+ \beta G_{ii}W_{ij} \geq \sum_k \left(X^T PP^T X\right)_{ik} w_{kj}^t \left(V^T CV\right)_{jj} \qquad , (37)$$

$$+ 2\alpha \sum_{i=1}^{N}\sum_k \left(X^T PP^T X\right)_{ik} w_{kj}^t (L_i)_{jj} + \beta \sum_k G_{ik} w_{kj}^t$$

$$\geq w_{ij}^t \left((K)_{ii}\left(V^T CV\right)_{jj} + 2\alpha \sum_{i=1}^{N}\left(X^T PP^T X\right)_{ii}(L_i)_{jj} + \beta G_{ii}\right)$$

$$\geq w_{ij}^t \frac{1}{2} F_{w_{ij}}''.$$

from which we can easily conclude that $\Delta\left(w, w_{ij}^t\right) \geq F_{w_{ij}}(w)$.

Similarly, the auxiliary function for the objective function with regard to the variable $v_{ij}$ is defined as follows:

**Lemma 3:** The following function

$$G\left(v, v_{ij}^t\right) = F_{v_{ij}}\left(v_{ij}^t\right) + F_{v_{ij}}'\left(v_{ij}^t\right)\left(v - v_{ij}^t\right)$$

$$+ \frac{\left(2CVW^T KW\right)_{ij} + \alpha(A+B)_{ij} + 2\beta GV_{ij}}{v_{ij}^t}\left(v - v_{ij}^t\right)^2 \qquad (38)$$

is an auxiliary function for $F_{v_{ij}}$, which is also the part of the objective function that is only relevant to $v_{ij}$.

**Proof:** This proof is essentially similar to that of Lemma 2. By comparing $\Delta\left(v, v_{ij}^t\right)$ with the Taylor series expansion of $F_{v_{ij}}$, we only need to prove $\left(2CVW^T KW\right)_{ij} + \alpha(A+B)_{ij} + 2\beta GV_{ij} / v_{ij}^t \geq 1/2F_{v_{ij}}''$. Note that we have $A, B \geq 0$ and

$$\left(2CVW^T KW\right)_{ij} + \alpha(A+B)_{ij} + 2\beta GV_{ij} \geq \left(2CVW^T KW\right)_{ij} + 2\beta GV_{ij}$$

$$= 2(CV)_{ij}\left(W^T KW\right)_{jj} + 2\beta G_{ii}V_{ij}$$

$$\geq 2\sum_k (C)_{ik} v_{kj}^t \left(W^T KW\right)_{jj} + 2\beta \sum_k G_{ik} v_{kj}^t \qquad .(39)$$

$$\geq v_{ij}^t \left(2(C)_{ii}\left(W^T KW\right)_{jj} + 2\beta G_{ii}\right) \geq v_{ij}^t \frac{1}{2} F_{v_{ij}}''.$$

Thus, we can similarly conclude that $\Delta\left(v, v_{ij}^t\right) \geq F_{v_{ij}}(v)$. In the similar ways, we can also obtain $\Delta\left(q, q_{ij}^t\right) \geq F_{q_{ij}}(q)$, and $\Delta\left(p, p_{ij}^t\right) \geq F_{p_{ij}}(p)$. Since $G\left(w, w_{ij}^t\right)$, $\Delta\left(v, v_{ij}^t\right)$, $\Delta\left(p, p_{ij}^t\right)$ and $\Delta\left(q, q_{ij}^t\right)$ are the auxiliary functions for $F_{w_{ij}}$, $F_{v_{ij}}$, $F_{p_{ij}}$ and $F_{q_{ij}}$, respectively. According to Lemma 1, by solving

$$w^{t+1} = \arg\min_w \Delta\left(w, w_{ij}^t\right), \quad v^{t+1} = \arg\min_v \Delta\left(v, v_{ij}^t\right),$$

$$p^{t+1} = \arg\min_p \Delta\left(p, p_{ij}^t\right), \quad q^{t+1} = \arg\min_q \Delta\left(q, q_{ij}^t\right),$$

we can obtain the same updates as in Eqs.(24-25)(28)(32) respectively. Thus, the optimization problem in Eq.(18) is non-increasing under these updates.

### 3.4 Computational Complexity Analysis

We present the computational complexity analysis of our RFA-LCF in this part. Note that we mainly describe the extra cost of RFA-LCF in comparison to LCF. We also use big O notation to show the complexity [28]. According to the updating rules in Eqs.(24-25), we can easily find that RFA-LCF has the same computational time complexity as LCF by using the big O when updating $W$ and $V$ specifically, i.e., $O(N^2 r)$, where $N$ is the number of samples in $X$ and $r$ is the dimension of new representation $V^T$. Besides $W$ and $V$, we also need to compute $P$ and $Q$ in each itera-tion. According to Eqs.(28)(32), the complexity of updating $P$ and $Q$ is at least $O(d^3)$, where $d$ is the dimension. Thus, the overall cost is $O(tN^2 r)$ when $N$ is far large than $d$ and the updates stop after $t$ times iterations.

## 4 RELATIONSHIP ANALYSIS

In this section, we discuss the connections between related methods and our RFA-LCF algorithm.

### 4.1 Connection with CF [5]

We first show that CF is a special case of our RFA-LCF. Recalling the objective function of our RFA-LCF in Eq.(18), if we constrain $\alpha = \beta = 0$, we have the following simplified formulation in the matrix trace expression:

$$\min_{W,V,P,b} tr\left(\left(X^T P + eb^T - VW^T X^T\right)^T M\left(X^T P + eb^T - VW^T X^T\right)\right) + \gamma \|P\|_{2,1}, \qquad (40)$$

where $M$ denotes a diagonal matrix with entries being $m_{ii} = 0.5 / \left\| \left(X^T P + eb^T - VW^T X^T\right)^i \right\|$. If the bias $b=0$, the projection is the standard basis (i.e., $P = I$) and $M$ is an identity matrix, we can obtain the following problem:

$$\min_{W,V} tr\left(\left(X^T - VW^T X^T\right)^T \left(X^T - VW^T X^T\right)\right) = \|X - XWV^T\|_F^2, \quad (41)$$

which is equivalent to the objective function of traditional CF problem in Eq.(1). But setting $P = I$ and $\Theta = I$ means that the factorization process will be performed on original data $X$, and the used metric will not be robust to noise and outliers in data any more. Setting the bias $b=0$ means that reconstruction error will lose the ability to handle the samples resided on a nonlinear manifold potentially. Setting $\alpha = \beta = 0$ means that the locality and sparsity between the anchor point $u_r$ and $x_i$ cannot be preserved any more, and the neighborhood information within the anchor points and learnt new representations cannot be encoded clearly. As a result, our RFA-LCF can potentially outperform CF for learning effective data representations.

### 4.2 Connection with LCCF [9]

We then discuss the relationship between our RFA-LCF and LCCF. Recalling the objective function of our RFA-LCF in Eq.(18), if we constrain $\alpha = \gamma = 0$, pre-calculate and fix the weight matrix $Q$ in the optimizations, the optimization problem in Eq.(18) can be reduced to

$$\min_{W,V,b} \|X^T P + eb^T - VW^T X^T\|_{2,1} + tr\left(\beta H^T GH\right)$$

$$= tr\left(\left(X^T P + eb^T - VW^T X^T\right) M\left(X^T P + eb^T - VW^T X^T\right)\right) \qquad (42)$$

$$+ tr\left(\beta H^T GH\right)$$

Suppose we further constrain $P = I$, fix $M$ to be an identity matrix, set $b=0$ and remove the neighborhood preservation constraint on $W$, the above problem becomes

$$\min_{W,V} \|X - XWV^T\|_F^2 + \left\|\begin{pmatrix}\sqrt{\beta}X \\ \sqrt{\beta}V^T\end{pmatrix} - \begin{pmatrix}\sqrt{\beta}X \\ \sqrt{\beta}V^T\end{pmatrix}Q\right\|_F^2. \qquad (43)$$

Note that $\|X - XQ\|_F^2$ is a constant when $Q$ is fixed in the optimization and the above problem is equivalent to LCCF if $G$ is equivalent to the graph Laplacian $L$ in LCCF. It should be noted that $G = L$ when $S = Q = ee^T / N$, where $e$ is a column vector of all ones. Compared with CF, LCCF suffers from the same shortcoming by setting $P=I$, $M=I$ and $\alpha=0$. In addition, fixing the weight matrix $Q$ by pre-calculating it before the reconstruction will clearly make LCCF lose the adaptive neighborhood preserving ability. Moreover, the pre-calculated weights cannot be ensured to be optimal for the subsequent factorization process.

### 4.3 Connection with LCF [10]

We discuss the relations between RFA-LCF and LCF. For the objective function of our RFA-LCF in Eq.(18), if we set $\beta=\gamma=0$, the problem in Eq.(18) can be reduced to

$$
\min_{W,V,b} tr\left(\left(X^T P+eb^T-VW^T X^T\right)M\left(X^T P+eb^T-VW^T X^T\right)\right) \\
+\alpha\sum_{i=1}^{N}\sum_{r=1}^{R}|v_{ri}|\left\|\sum_{j=1}^{N}w_{jr}P^T x_j-P^T x_i\right\|^2 \tag{44}
$$

which is equivalent to the formulation of LCF if $M=I$, $P=I$ and $b=0$. But setting $P=I$, i.e., standard basis, the locality and sparsity of anchor points and corresponding coordinates are encoded in the original data space, which may result in the inaccurate measures of the locality and sparsity due to the possibly included noise in data.

### 4.4 Connection with GRLCF [26] and LGCF [48]

Both GRLCF and LGCF are the combination of LCCF and LCF by considering the manifold structures of data and the locality as an additional constraint simultaneously by solving the following minimization problem:

$$
\min_{W,V}\|X-XWV\|^2+\lambda\sum_{i=1}^{N}\sum_{r=1}^{R}|v_{ri}|\left\|\sum_{j=1}^{N}w_{jr}x_j-x_i\right\|^2+\mu tr\left(VLV^T\right), \tag{45}
$$

where $\lambda,\mu\geq 0$, and $L=D-S$, where $D$ is a diagonal matrix whose entries are column (or row) sums of the weight matrix $S$. The only difference between GRLCF and LGCF is that they define the graph weights in two differernt ways. Specifically, LGCF defines the weight matrix by using the cosine similarity based on the $k$-nearest-neighbor adjacency graph, while GRLCF encodes the graph weights with exactly $p$ connected components (where $p$ is the number of clusters) by a novel graph-based CLR algorithm [18], and then defines $S$ by solving the following problem:

$$
\min_{S}\|S-S_0\|^2,\ \ s.t.\ \sum_{j}S_{ij}=1,\ S_{ij}\geq 0,\ rank\left(L\right)=N-p, \tag{46}
$$

where $S_0$ is an initial weight matrix, $L=E-(S^T+S)/2$, $E$ is a diagonal matrix whose entries are given by $E_{ii}=\sum_{j}S_{ij}$ and $N$ is the number of samples. If we constrain $\gamma=0$ in our RFA-LCF, the objective function in Eq.(18) can be rewritten as

$$
\min_{W,V,Q,M,b} tr\left(\left(X^T P+eb^T-VW^T X^T\right)M\left(X^T P+eb^T-VW^T X^T\right)\right) \\
+\alpha\sum_{i=1}^{N}\sum_{r=1}^{R}|v_{ri}^T|\left\|\sum_{j=1}^{N}w_{jr}P^T x_j-P^T x_i\right\|+\beta tr\left(H^T GH\right). \tag{47}
$$

Note that Eq.(47) is equal to Eq.(45) when the following conditions are satisfied: (1) if the diagonal matrix $M$ is set to an identity matrix; (2) projection $P$ is the standard basis (i.e., $P=I$); (3) the bias vector $b=0$; (4) $G$ is equivalent to the Laplacian matrix $L=D-S$ and the neighborhood preserving constraint on the basis vectors $W$ and embedded features $P^T X$ are removed. That is, the problem in Eq.(45) is just a reduced formulation of RFA-LCF in Eq.(47).

## 5 SIMULATION RESULTS AND ANALYSIS

We conduct simulations to examine our RFA-LCF for data clustering and representation. The results of our RFA-LCF are compared with those of 12 related algorithms, i.e., NMF [4], PNMF [6], GNMF [7], DNMF [41], DSNMF [49], PAMGNMF [50], CF [5], LCCF [9], LCF [10], LGCF [48], GRLCF [26] and GCF [40], which are all closely related to our algorithm. Note that there are no parameters in NMF, PNMF and CF, and the hyperparameters of GNMF, DNMF, DSNMF, LCF, LCCF, GRLCF, LGCF, PAMGNMF and GCF are carefully chosen for fair comparison. Eight public databases are evaluated, including three face databases (i.e., ORL [15], UMIST [33] and CMU PIE [34]), two object databases (ETH80 [32] and COIL100 [13]), two handwritten datasets from CASIA-HWDB1.1 [12] and one time series database (SCC) [61]. For face and object databases, images are resized into 32×32 pixels (i.e., each image is represented using a 1024-dimensional vector). Detailed information about used datasets are shown in Table 1. We perform all experiments on a PC with Intel Core i5-4590 CPU @ 3.30 GHz 3.30 GHz 8G.

TABLE 1: LIST OF USED DATASETS AND DATASET INFORMATION.

| Data Type | Dataset Name | # Points | # Dim | # Class |
|---|---|---|---|---|
| Face images | ORL [15] | 400 | 1024 | 40 |
| | UMIST [21] | 1012 | 1024 | 20 |
| | CMU PIE [34] | 11554 | 1024 | 68 |
| Object images | ETH80 [32] | 3280 | 1024 | 80 |
| | COIL100 [13] | 7200 | 1024 | 100 |
| Handwritten images | HWDB1.1-D [12] | 2381 | 196 | 10 |
| | HWDB1.1-L [12] | 12456 | 256 | 52 |
| Time Series | SCC [61] | 600 | 60 | 6 |

### 5.1 Quantitative Clustering Evaluation

**Clustering Evaluation Process.** For the quantitative clustering evaluations, we perform K-means with cosine distance on the new representations obtained by each method. Following the procedures in [19][23], for each number K of clusters, we choose K categories from each set randomly and use the data of K categories to form the matrix $X$ as [19][23]. For each algorithm, the rank $r$ is set to K+1 as [19] and we average the numerical results over 30 random initializations for the K-means clustering algorithm.

**Clustering Evaluation Metric.** We employ two widely-used quantitative evaluation metrics, i.e., *Accuracy* (AC) and *F-measure* [22][24]. AC is the percentage of the cluster labels to the true labels provided by the original data corpus. More specifically, AC is defined as follows:

$$
AC=\frac{\sum_{i=1}^{N}\delta\left(r_i,map\left(p_i\right)\right)}{N}, \tag{48}
$$

where $N$ is the number of samples, and the map function $map(p_i)$ is the permutation mapping function that maps the cluster label $p_i$ obtained by clustering to the true label $r_i$ provided by the data corpus, and the best mapping solution is obtained by the Kuhn-Munkres algorithm [55] according to [56]. More specifically, $\delta(r_i,map(p_i))=1$ when $r_i=map(p_i)$, and else $\delta(r_i,map(p_i))=0$.

The defitnniion of the F-measure is presented as follows:

$$
F_\mu=\frac{\left(\mu^2+1\right)\text{Precision}\times\text{Recall}}{\mu^2\text{Precision}+\text{Recall}}, \tag{49}
$$

where we set the parameter $\mu = 1$. Note that both values of the AC and F-measure range from 0 to 1, i.e., the higher the value is, the better the clustering result will be.

## 5.2 Visualization of Graph Adjacency Matrix

We compare the adaptive reconstruction weight matrix $Q$ of our RFA-LCF with the binary weights used in GCF and DNMF, Cosine similarity weights used in LCCF, and the CLR weights [18] used in GRLCF. ORL database is used as an example. For clear observation, we choose images of 10 people to construct the adjacency graphs. The nearest neighbor number is set to 7 [19] for each approach for fair comparison. To evaluate the robustness of the weighting to noise, we also prepare a setting under noisy case.

We visualize the constructed weight matrices on original and noisy data in Fig.1, respectively. To corrupt data, we add random Gaussian noise to original data $X$, where the variance is fixed to 20. We evaluate the graph adjacency matrices numerically using the reconstruction error $\varepsilon = \left\| X - XQ \right\|_F^2 / \left\| X \right\|_F^2$, where $Q$ is a weight matrix obtained by each weighting approach. Clearly, the smaller the reconstruction error is, the better the reconstruction result will be, and vice versa. We can find that: 1) the constructed weight matrices by each weighting method have approximate block-diagonal structures; 2) more noisy or wrong inter-class connections are produced in the binary, Cosine and CLR weights, which may potentially lead to inaccurate similarity measures and high clustering error, compared with our adaptive weights; 3) the reconstruction error of each method over the noisy data is higher than that on original data, which implies that the noise in data can indeed decrease the representation ability of encoded weights; 4) our adaptive weight matrix can deliver smaller reconstruction errors than other weighting methods in both original and noisy cases, i.e., using our adaptive weight matrix to reconstruct data is potentially more accurate; 5) the reconstruction error by CLR weights is larger than that by cosine similarity weights and binary weights in both original and noisy cases.

## 5.3 Convergence Analysis

The problem of our RFA-LCF is solved alternately, so we present its convergence analysis results on two databases (i.e., COIL100 and HWDB1.1-D). The convergence analysis results are shown in Figs.2-3, where the convergence results of closely related LCF, LCCF and GRLCF are also provided for comparison. We mainly illustrate the divergence between two consecutive new representations V of each method for fair comparison. We can find that: 1) the divergence between two consecutive new representations of each algorithm is non-increasing; 2) all the factorization methods have a relatively rapid convergence rate and the number of iterations is about 20; 3) the convergence speed of our RFA-LCF is comparable to other related methods.

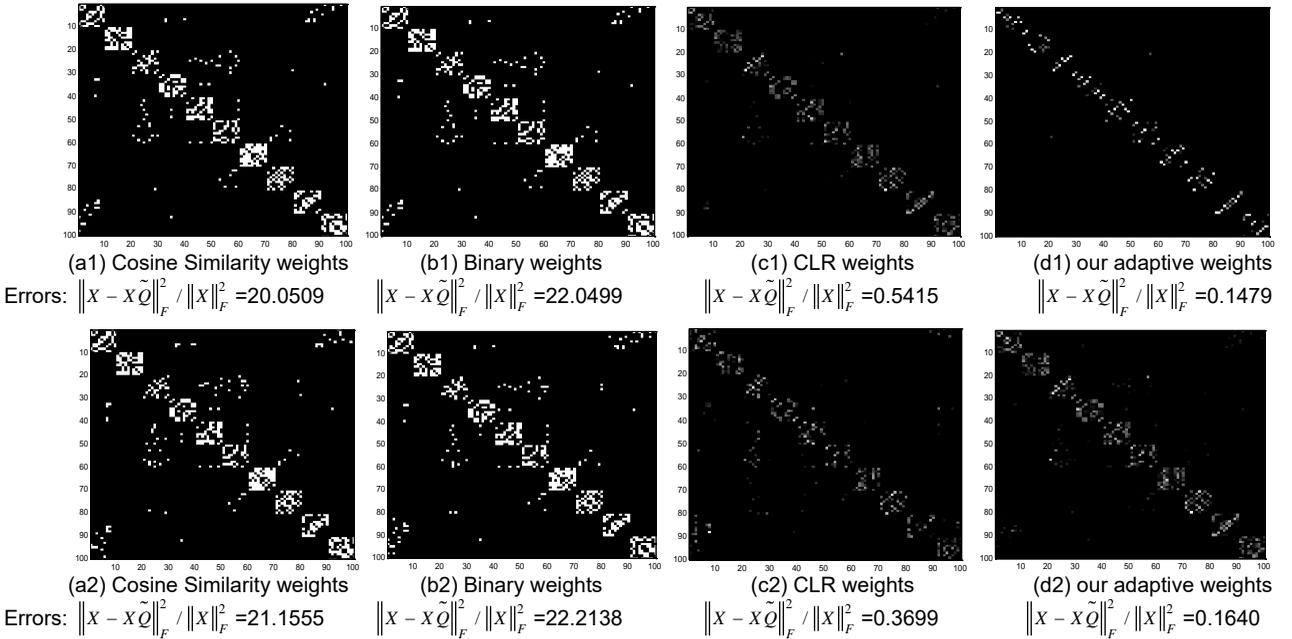

| (a1) Cosine Similarity weights | (b1) Binary weights | (c1) CLR weights | (d1) our adaptive weights |

Errors: $\left\| X - X\tilde{Q} \right\|_F^2 / \left\| X \right\|_F^2 = 20.0509$    $\left\| X - X\tilde{Q} \right\|_F^2 / \left\| X \right\|_F^2 = 22.0499$    $\left\| X - X\tilde{Q} \right\|_F^2 / \left\| X \right\|_F^2 = 0.5415$    $\left\| X - X\tilde{Q} \right\|_F^2 / \left\| X \right\|_F^2 = 0.1479$

| (a2) Cosine Similarity weights | (b2) Binary weights | (c2) CLR weights | (d2) our adaptive weights |

Errors: $\left\| X - X\tilde{Q} \right\|_F^2 / \left\| X \right\|_F^2 = 21.1555$    $\left\| X - X\tilde{Q} \right\|_F^2 / \left\| X \right\|_F^2 = 22.2138$    $\left\| X - X\tilde{Q} \right\|_F^2 / \left\| X \right\|_F^2 = 0.3699$    $\left\| X - X\tilde{Q} \right\|_F^2 / \left\| X \right\|_F^2 = 0.1640$

Fig.1: Visualization comparison of the constructed weights by each weighting approach on the ORL face database, where (a1)-(d1) are based on the original data and (a2)-(d2) are based on the noisy data with corruptions.

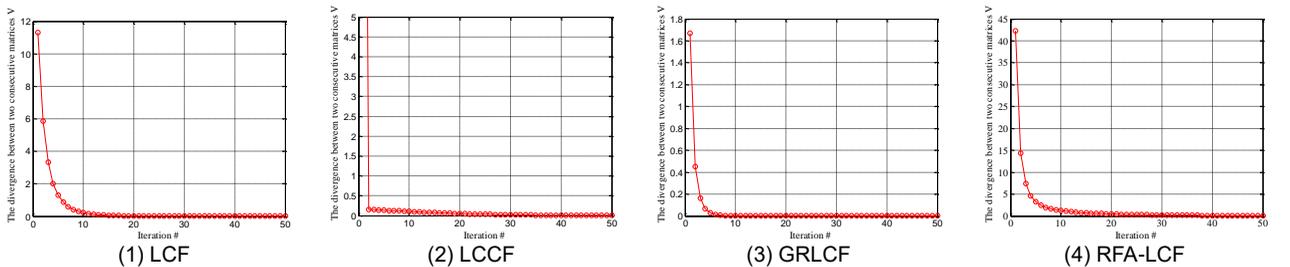

| (1) LCF | (2) LCCF | (3) GRLCF | (4) RFA-LCF |

Fig.2: Convergence curve of each evaluated method on the COIL100 object database.

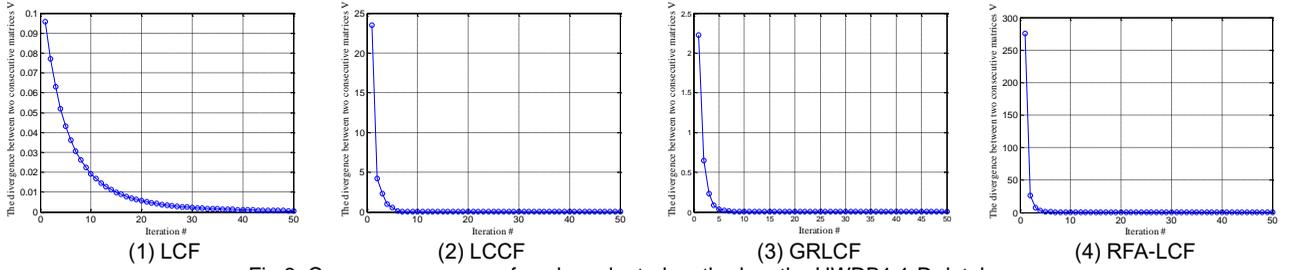

(1) LCF    (2) LCCF    (3) GRLCF    (4) RFA-LCF

Fig.3: Convergence curve of each evaluated method on the HWDB1.1-D database.

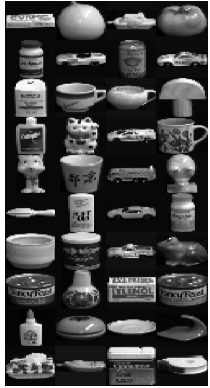 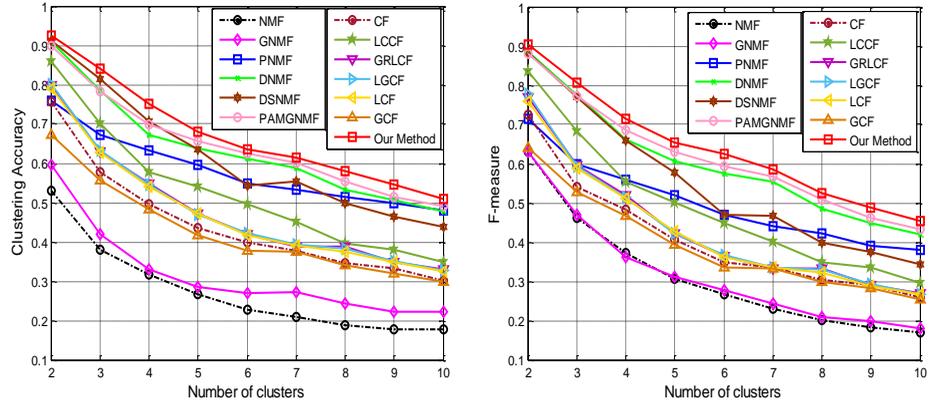

Fig.4: Sample images and quantitative clustering evaluation results on the COIL100 object database.

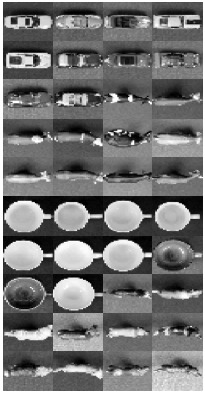 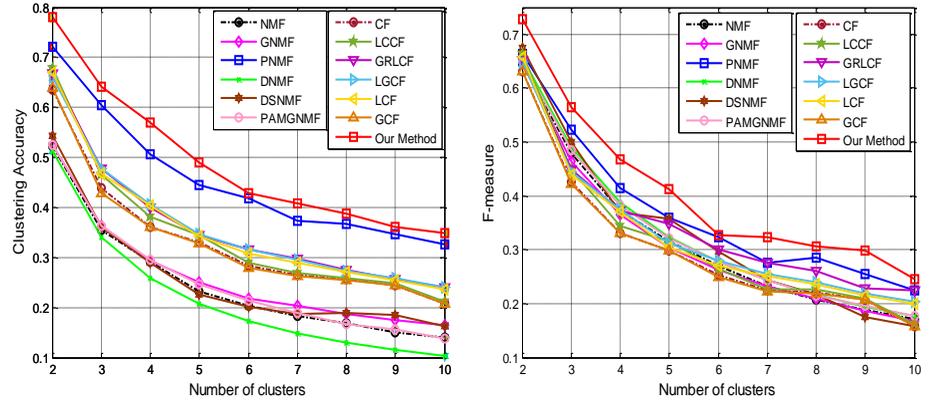

Fig.5: Sample images and quantitative clustering evaluation results on the ETH80 object database.

## 5.4 Object Clustering

We first evaluate each algorithm for clustering the object images of COIL100 and ETH80 databases. Based on the clustering evaluation process in Subsection 5.1, we average the results over 10 random selections of the K categories to avoid the contingency. The clustering curves of AC and F-measure are illustrated in Figs.4-5. We can find that: (1) the delivered AC and F-measure of each algorithm go down as the number of categories increases, since clustering data of less categories is relatively easy; (2) our RFA-LCF method delivers higher values of AC and F-measure than other methods in the investigated cases. PAMGNMF and DNMF also work well by obtaining higher accuracies than other compared methods on the COIL100 database. PNMF also works well on the ETH80 database by delivering higher accuracies than other compared methods.

## 5.5 Face Clustering

We then evaluate each method for clustering the face image data of the CMU PIE and UMIST face databases. In this study, the random features of face images are applied. To extract random features, each image is projected onto a $d$-dimensional feature vector with a randomly generated matrix from a zero-mean normal distribution. Each row of the matrix is $L_2$-normalized. Similar to [35-37], the dimensionality of random face image features is set to $d$=540 in our study. We perform clustering over the randomly extracted data of K categories. To avoid the randomness by bias and to achieve a fair comparison, the accuracy and F-measure are averaged over 10 random selections of the K-category data. The clustering results on CMU PIE and UMIST databases are illustrated in Fig.6 and Fig.7. We can find that: (1) the AC and F-measure of each algorithm decrease with the increasing numbers of categories; (2) our RFA-LCF can deliver higher AC and F-measure than the other methods in most cases; (3) CF and NMF usually obtain the worst result on each database.

## 5.6 Handwriting Clustering

In this study, two handwriting datasets are used to evalu-ate each method, i.e., HWDB1.1-D and HWDB1.1-L that are two sample sets of the CASIA-HWDB1.1 handwriting database [12]. Specifically, HWDB1.1-D includes 2381 handwritten digits ('0'-'9') of 14×14 pixels and HWDB1.1-L includes 12456 handwritten letters ('a'-'z' and 'A'-'Z') of 16×16 pixels from 52 classes. The averaged AC and F-measure on HWDB1.1-D and HWDB1.1-L are shown in Figs.8 and 9. We can find that: (1) the increasing number of categories can decrease the AC and F-measure of each model due to the fact that clustering data of less categories is relatively easier; (2) RFA-LCF delivers higher AC and F-measure than other related methods in most cases; (3) CF and NMF still obtain the worst result on each set.

The statistics (i.e., the mean and best records over AC) according to Figs.4-9 are reported in Table 2. Similar performance superiority of the methods can be concluded.

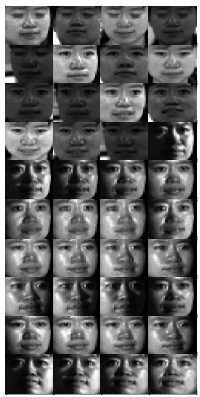
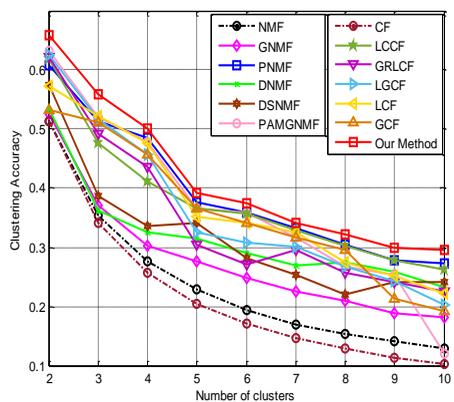
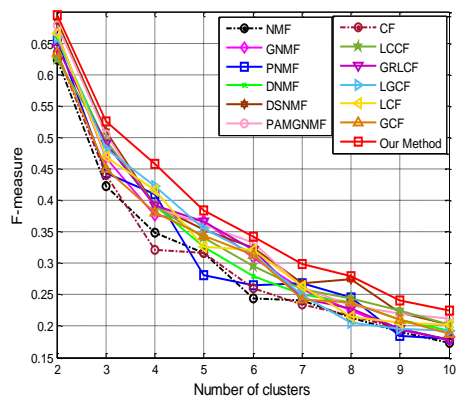

Fig.6: Sample images and quantitative clustering evaluation results on the CMU PIE face database.

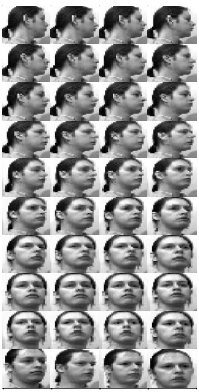
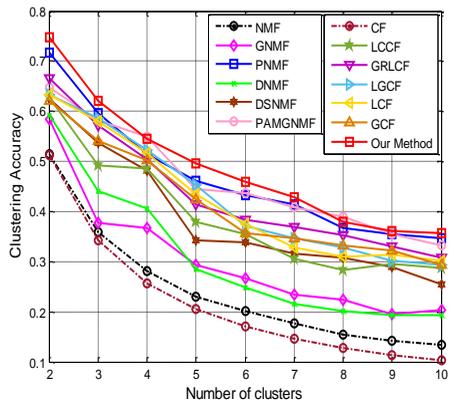
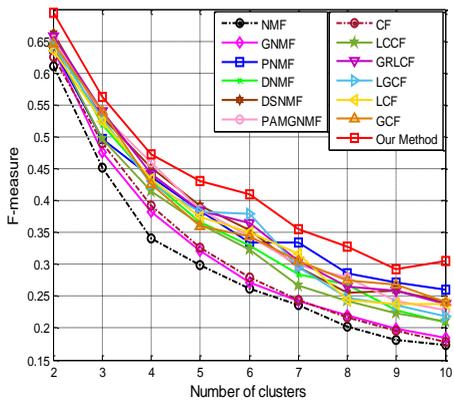

Fig.7: Sample images and quantitative clustering evaluation results on the UMIST face database.

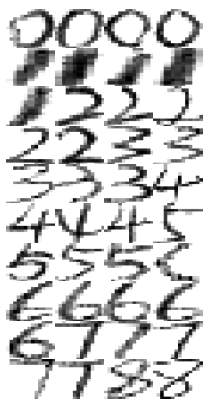
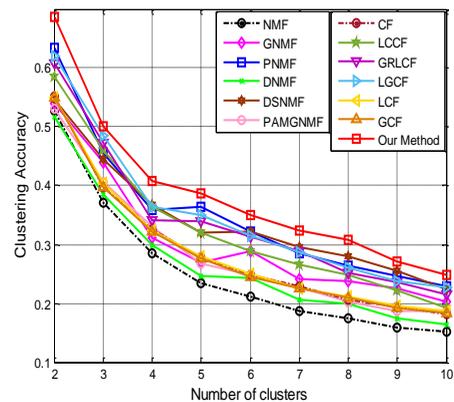
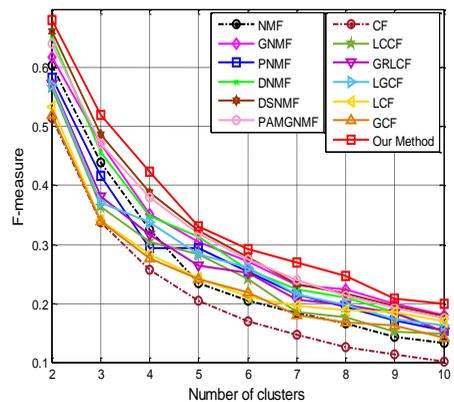

Fig.8: Sample images and quantitative clustering evaluation results on the HWDB1.1-D database.

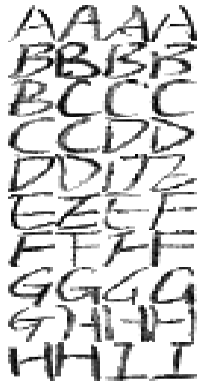
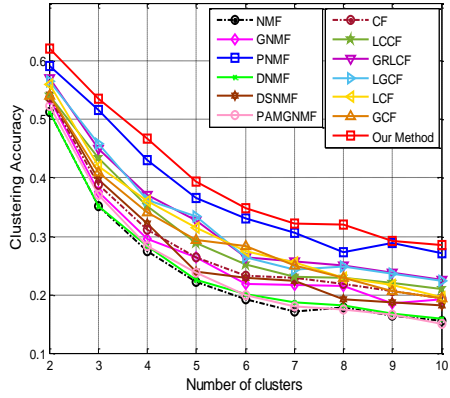
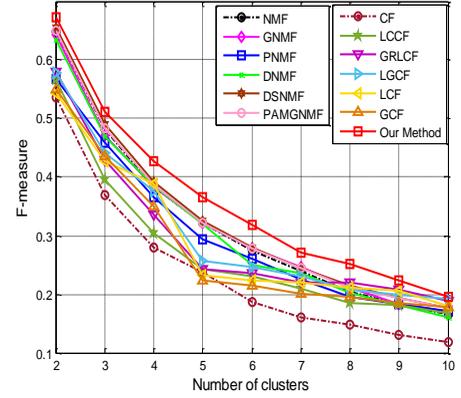

Fig.9: Sample images and quantitative clustering evaluation results on the HWDB1.1-L database.

TABLE 2: MEAN AND HIGHEST CLUSTERING ACCURACY (AC) ON FACE, OBJECT AND HANDWRITING DATABASES.

| Dataset / Method | CMU PIE face | | UMIST face | | COIL100 object | |
|---|---|---|---|---|---|---|
| | Mean±std (%) | Best (%) | Mean±std (%) | Best (%) | Mean±std (%) | Best (%) |
| NMF | 23.99±12.51 | 51.2 | 24.36±12.49 | 51.4 | 27.50±11.76 | 52.9 |
| GNMF | 28.10±10.96 | 52.6 | 30.50±12.39 | 58.5 | 31.84±12.09 | 59.6 |
| PNMF | 39.20±11.73 | 60.8 | 46.72±12.34 | 71.6 | 58.14±9.25 | 76.0 |
| DNMF | 31.82±9.01 | 53.5 | 30.86±14.08 | 59.3 | 63.63±13.86 | 90.9 |
| DSNMF | 31.98±11.08 | 57.5 | 38.83±12.78 | 62.5 | 61.85±16.29 | 91.1 |
| PAMGNMF | 36.81±15.53 | 63.3 | 46.04±10.86 | 64.7 | 64.71±13.14 | 89.9 |
| CF | 21.97±13.38 | 51.3 | 21.97±13.38 | 51.3 | 44.61±14.44 | 75.5 |
| LCCF | 37.80±11.25 | 61.9 | 39.01±11.96 | 62.6 | 52.78±16.58 | 85.8 |
| GRLCF | 34.94±13.65 | 62.3 | 43.29±12.14 | 66.5 | 48.03±15.33 | 79.8 |
| LGCF | 36.06±14.07 | 62.5 | 42.61±12.78 | 63.3 | 48.05±15.48 | 80.2 |
| LCF | 37.09±12.41 | 57.3 | 42.15±12.66 | 63.3 | 47.61±15.27 | 79.1 |
| GCF | 35.82±12.14 | 53.3 | 41.57±11.42 | 61.1 | 42.71±12.31 | 67.3 |
| **RFA-LCF** | **41.58±12.84** | **65.9** | **48.83±13.10** | **74.7** | **67.59±13.88** | **92.5** |

| | ETH80 object | | HWDB1.1-D handwriting | | HWDB1.1-L handwriting | |
|---|---|---|---|---|---|---|
| | Mean±std (%) | Best (%) | Mean±std (%) | Best (%) | Mean±std (%) | Best (%) |
| NMF | 24.94±12.42 | 52.3 | 25.55±12.34 | 52.7 | 24.67±11.84 | 51.3 |
| GNMF | 26.42±11.60 | 52.4 | 30.64±11.18 | 54.0 | 27.74±11.40 | 53.6 |
| PNMF | 45.65±13.24 | 72.1 | 35.08±12.73 | 63.4 | 37.51±11.52 | 59.2 |
| DNMF | 22.06±13.33 | 51.2 | 27.02±11.49 | 51.7 | 25.16±11.59 | 51.3 |
| DSNMF | 26.06±12.27 | 54.2 | 33.95±10.10 | 54.7 | 27.93±12.15 | 54.3 |
| PAMGNMF | 25.49±12.41 | 52.5 | 28.73±11.63 | 53.1 | 25.39±12.33 | 52.4 |
| CF | 33.65±13.23 | 63.5 | 28.99±12.01 | 55.3 | 28.66±11.21 | 53.7 |
| LCCF | 34.97±14.58 | 67.9 | 32.74±12.58 | 58.5 | 30.62±11.47 | 54.1 |
| GRLCF | 36.39±13.61 | 66.8 | 34.07±12.59 | 60.8 | 32.79±11.72 | 57.1 |
| LGCF | 36.34±13.42 | 66.0 | 34.95±12.88 | 62.1 | 32.63±11.83 | 56.6 |
| LCF | 36.04±13.79 | 67.1 | 29.17±11.91 | 54.9 | 31.43±11.74 | 56.1 |
| GCF | 33.40±13.29 | 64.0 | 28.89±11.97 | 55.0 | 30.52±11.16 | 54.2 |
| **RFA-LCF** | **49.05±14.63** | **78.0** | **38.71±13.59** | **68.7** | **39.85±11.83** | **62.2** |

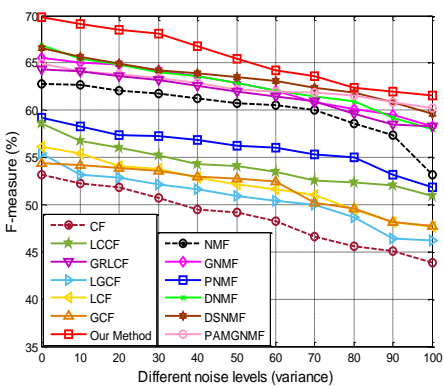
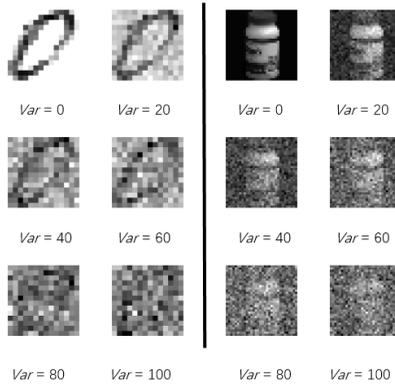
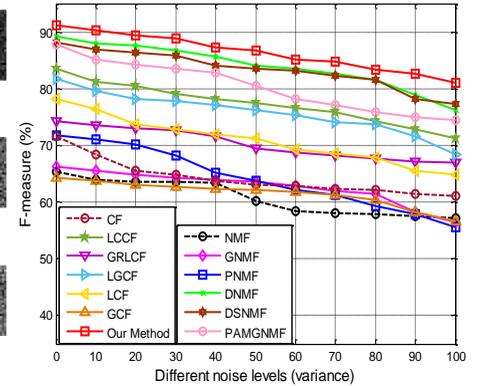

(a) HWDB1.1-D    (b) Image examples with corrupted pixels    (c) COIL100

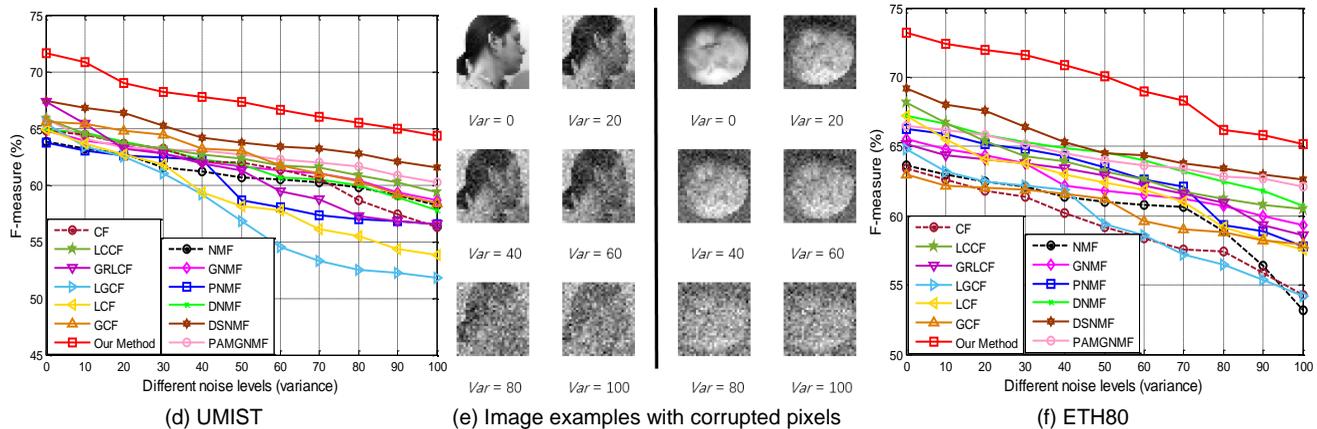

Fig.10: Clustering image data of (a) HWDB1.1-D, (c) COIL100, (d) UMIST and (f) ETH80 against different levels of corruptions.

*Image labels within figure (e):* Var = 0, Var = 20, Var = 40, Var = 60, Var = 80, Var = 100; (d) UMIST; (e) Image examples with corrupted pixels; (f) ETH80

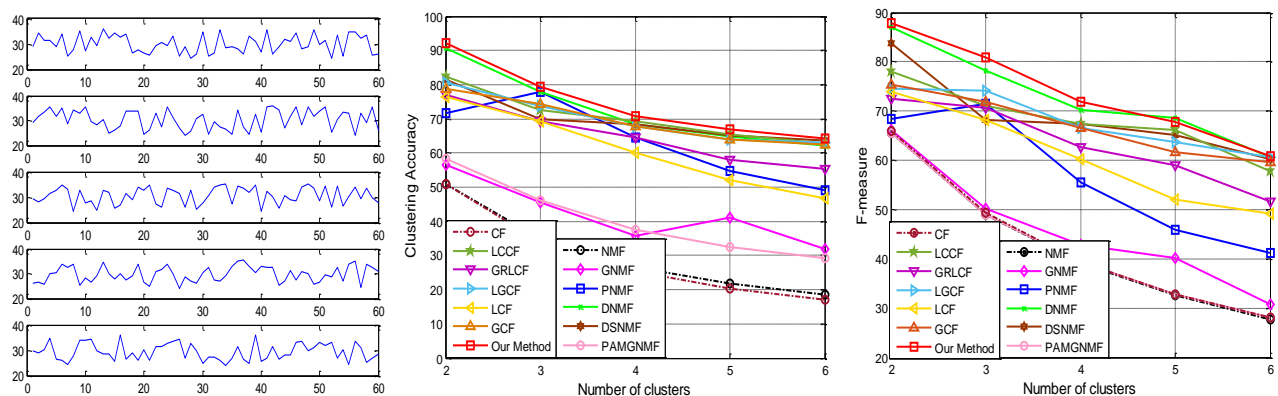

Fig.11: Visualized samples and quantitative clustering evaluation results on the synthetic control chart time series dataset.

## 5.7 Clustering Image Data against Corruptions

In addition to evaluating each algorithm for clustering the original images, we also conduct simulations to test them for clustering the noisy image data. In this study, four real image databases, i.e., UMIST, ETH80, HWDB1.1-D and COIL100, are evaluated. To corrupt the data matrix $X$, we add random Gaussian noise with the variance being 0-100 with interval 10 into the selected pixles of images. Note that the position of the corrupted pixles is unknown to users. Fig. 10 shows the corrupted images and the clustering results, by F-measure, with different levels of noise. The results of F-measure are obtained by randomly choosing two categories each time, and averaging over 50 runs of k-means clustering to avoid the randomness. From the results, we can find that: (1) the clustering result of each method generally goes down with the increasing levels of noise, i.e., the corrupted noisy data can indeed cause negative effects on the data representation and clustering; (2) our RFA-LCF can outperform other methods in this study, since our RFA-LCF clearly incorporates the error correction procedure into the representation learning, and the factorization, sparse local-coordinate coding, and reconstruction error are defined over the noise-removed clean data. This finding can also demonstrate that our proposed robust learning strategy is feasible and effective. DNMF also works well on clustering the object images of ETH80 and COIL100, while LCCF works well on clustering the face images of UMIST and object images of ETH80. The results of DNMF and LCCF are comparable on ETH80.

Table 3: Mean and Highest Clustering Accuracy (AC) on Synthetic Control Chart Time Series Dataset.

| Dataset | Time series dataset (SCC) | |
|---|---|---|
| Method | Mean±std (%) | Best (%) |
| NMF | 30.54±12.96 | 51.0 |
| GNMF | 42.15±9.51 | 56.4 |
| PNMF | 63.60±11.78 | 77.9 |
| DNMF | 72.79±11.55 | 90.5 |
| DSNMF | 69.73±6.97 | 81.4 |
| PAMGNMF | 40.71±11.79 | 58.4 |
| CF | 29.48±13.54 | 50.8 |
| LCCF | 70.45±7.58 | 82.4 |
| GRLCF | 64.87±8.79 | 77.1 |
| LGCF | 69.78±7.40 | 80.7 |
| LCF | 60.89±12.14 | 76.4 |
| GCF | 69.41±7.01 | 78.9 |
| **RFA-LCF** | **74.64±11.30** | **92.0** |

## 5.8 Clustering Time Series Data

In this section, we also investigate the clustering ability of our RFA-LCF method for handling the time series dataset. A standard UCI dataset, i.e., synthetic control chart time series dataset (SCC) [61], is evaluated in this study, which contains examples of six different classes of control chart time series. We mainly vary the number K from {2, 3, ..., 6} and calculate the clustering accuracy and F-measure. To avoid the randomness, the results of clustering accuracy and F-measure are averaged over 30 random selections of the K categories data for each evaluated algorithm for fair comparison. The clustering evaluation results are shown in Fig.11, and the averaged results according to Fig.11 are given in Table 3. We find that: 1) our proposed method obtains enhanced performance in most cases; 2) besides our model, DNMF and DSNMF can usually obtain higher clustering results than other evaluated methods.

### 5.9 Parameter Sensitivity Analysis

***Investigation of the parameters of our RFA-LCF.*** We first investigate the effects of model parameters $\alpha$, $\beta$ and $\gamma$ on the clustering results of our RFA-LCF. In this study, F-measure is used as the quantitative evaluation metric and UMIST face database is used as an example. Since there are three parameters in RFA-LCF, we adopt the widely-used grid search strategy [20-21], i.e., fixing one of the parameters and tuning other two from $\left\{10^{-8}, 10^{-6}, \ldots, 10^{8}\right\}$. The settings for the parameter analysis are shown below: (1) we first fix $\gamma = 1$ to tune $\alpha$ and $\beta$; (2) we then fix $\alpha = 10^4$ and vary $\beta$ and $\gamma$; (3) finally, we fix $\beta = 10^6$ and vary $\alpha$ and $\gamma$. The analysis results are illustrated in Fig.11, where the number of categories is set to two, and the results are averaged over 30 random selections of categories and the central points in K-means clustering. We find that RFA-LCF delivers stable results over a wide range of parameter settings, i.e., our method is robust to the parameters. As a result, selecting proper parameters for RFA-LCF will be relatively easy in practical applications. Based on the parameter analysis results, we simply set $\alpha = 10^4$, $\beta = 10^6$ and $\gamma = 10^{-4}$ for RFA-LCF in the simulations.

In addition to presenting the above parameter analysis, we also explore the effects of the three terms involved in the objective function by respectively setting $\alpha = 0$, $\beta = 0$ and $\gamma = 0$. In this study, the number of categories is also set to two for clustering, and the results are averaged over 30 random selections of categories. The clustering results are described in Table 4. Note that $\alpha$ trades off the robust adaptive local-coordinate coding, $\beta$ trades off the adaptive weight learning and $\gamma$ trades off the robust projection learning for recovery. From the results, we can find that: 1) setting $\alpha = 0$ has resulted in the most decreased clustering results in most cases, implying that the robust adaptive local-coordinate coding seems to have the most significant effect on the performance; 2) by setting $\beta = 0$ or $\gamma = 0$, the clustering results also drop by 5%-25%. As a result, the three terms are all important in improving the performance of our proposed algorithm.

***Investigation of the hyper-parameters of other competitors.*** We also present the hyper-parameter analysis of the other compared methods and report the used parameters. But because of the page limitation, we have presented the detailed parameter analysis results of the other competitors in the supplementary document.

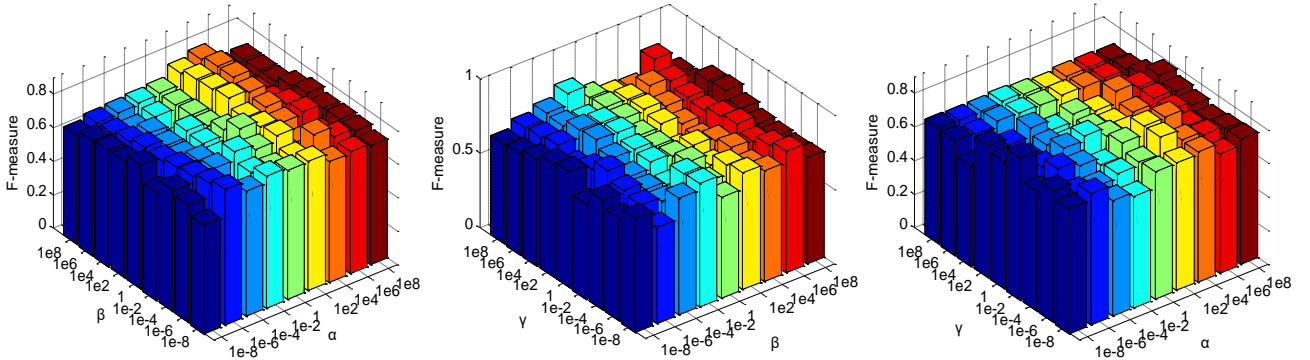

Fig.11: Clustering performance of our RFA-LCF under various parameters on the UMIST face database, where: *(left)* Fix $\gamma$ to tune $\alpha$ and $\beta$; *(middle)* Fix $\alpha$ to tune $\beta$ and $\gamma$; *(right)* Fix $\beta$ to tune $\alpha$ and $\gamma$.

TABLE 4: CLUSTERING ACCURACY (AC) OVER DIFFERENT PARAMETER SETTINGS ON USED DATABASES.

| Parameter settings | CMU PIE | UMIST | COIL100 | ETH80 | HWDB1.1-D | HWDB1.1-L | SCC |
|---|---|---|---|---|---|---|---|
| $\alpha = 10^4, \beta = 10^6, \gamma = 10^{-4}$ | **65.9** | **74.7** | **92.5** | **78.0** | **68.7** | **62.2** | **92.0** |
| $\alpha = 0, \beta = 10^6, \gamma = 10^{-4}$ | 52.5 | 53.7 | 51.4 | 51.6 | 51.0 | 51.0 | 50.9 |
| $\alpha = 10^4, \beta = 0, \gamma = 10^{-4}$ | 55.0 | 66.9 | 73.7 | 67.1 | 58.3 | 57.9 | 76.8 |
| $\alpha = 10^4, \beta = 10^6, \gamma = 0$ | 59.6 | 65.6 | 67.4 | 67.6 | 51.0 | 55.8 | 77.3 |

## 6 CONCLUDING REMARKS

We proposed a novel and effective robust flexible auto-weighted local-coordinate concept factorization model for unsupervised representation and clustering. Our framework aims to improve the accuracy of the data representation and encoded neighborhood against noise and corruptions by seamlessly integrating the robust flexible CF, robust sparse local coordinate coding, error correction, and adaptive reconstruction weight learning. The applied $L_{2,1}$-norm based flexible reconstruction residue can minimize the factorization error. The used adaptive weighting strategy can also avoid the tricky process of selecting the optimal parameters in defining the affinity, which is suffered in existing CF variants. Moreover, the flexible residue, local-coordinate coding and adaptive weight learning are all jointly performed based on the recovered clean data space by error correction, which can potentially lead to enchanced representations for clustering.

We have evaluated our RFA-LCF method for clustering three kinds of data. The numerical results have demonstrated the effectiveness of RFA-LCF for representing and grouping data by comparing with 12 related factorization methods. Although promising results are delivered, several challenging tasks related to our algorithm should be addressed in future. First, how to evaluate RFA-LCF to involve new data remains unclear, and thus the inductive extension will be explored. Second, extending our method to classification and retrieval can be explored. Third, selecting the optimal rank of factorization is still an open issue in CF based models, which should also be studied.

## ACKNOWLEDGEMENTS

We want to express our sincere gratitude to anonymous referee and their comments that make our manuscript a higher standard. This work is partially supported by the National Natural Science Foundation of China (61672365, 61732008, 61725203, 61622305, 61871444, 61572339), High-Level Talent of "Six Talent Peak" Project of Jiangsu Province (XYDXX-055), and the Fundamental Research Funds for the Central Universities of China (JZ2019HGPA0102). Zhao Zhang is the corresponding author of this paper.

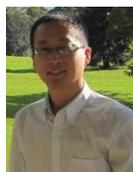

**Zhao Zhang** (SM'17- ) received the Ph.D. degree from the Department of Electronic Engineering (EE), City University of Hong Kong, in 2013. He is now a Full Professor at the School of Computer Science & School of Artificial Intelligence, Hefei University of Technology, Hefei, China. Dr. Zhang was a Visiting Research Engineer at National University of Singapore from Feb to May 2012. He then visited the National Laboratory of Pattern Recognition (NLPR) at Chinese Academy of Sciences from Sep to Dec 2012. During Oct 2013 and Oct 2018, he was an Associate Professor at the School of Computer Science and Technology, Soochow University, Suzhou, China. His current research interests include Multimedia Data Mining & Machine Learning, Image Processing & Pattern Recognition. He has authored/co-authored over 80 technical papers published at prestigious journals and conferences, such as IEEE TKDE (5), IEEE TIP (4), IEEE TNNLS (4), IEEE TCSVT, IEEE TCYB, IEEE TSP, IEEE TBD, IEEE TII (2), ACM TIST, Pattern Recognition (6), Neural Networks (8), Computer Vision and Image Understanding, Neurocomputing (3), IJCAI, ACM Multimedia, ICDM, ICASSP and ICMR, etc. Specifically, he has published 17 regular papers in the IEEE/ACM Transactions journals as the first-author/corresponding author. Dr. Zhang is serving/served as an Associate Editor (AE) for IEEE Access, Neurocomputing and IET Image Processing. Besides, he has been acting as a Senior Program Committee (SPC) member or Area Chair (AC) of PAKDD、BMVC、ICTAI, and a PC member for 10+ popular international conferences. He is now a Senior Member of the IEEE.

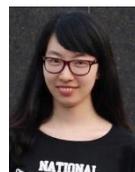

**Yan Zhang** is now working toward the PHD degree at the School of Computer Science and Technology, Soochow University, Suzhou, China, co-supervised by Dr. Zhao Zhang. Her current research interests mainly include data mining, machine learning and pattern recognition. Specifically, she is interested in high-dimensional data analysis and feature extraction. During her PhD study, she has already published papers in the IEEE TKDE and Pattern Recognition.

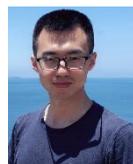

**Sheng Li** (SM'19- ) received the Ph.D. degree from Northeastern University, Boston, MA, in 2017. He is currently a Tenure-Track Assistant Professor in the Department of Computer Science at University of Georgia. From 2017-2018, heworked as a Data Scientist at the Big Data Experience Lab, Adobe Research, San Jose, CA. He has published over 50 papers at leading conferences and journals. He received the best paper awards (or nominations) at SDM 2014, IEEE ICME 2014, and IEEE FG 2013. He serves on the Editorial Board of Neural Computing and Applications, Neurocomputing, and serves as an Associate Editor of IEEE Computational Intelligence Magazine, Neurocomputing and IET Image Processing, etc. He has also served as a reviewer for several IEEE Transactions, and program committee member for NIPS, IJCAI, AAAI, and KDD. His research interests include robust machine learning, dictionary learning, visual intelligence, and behavior modeling. He is now a Senior Member of the IEEE.

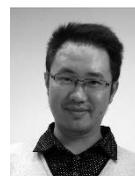

**Guangcan Liu** (M'11-SM'18- ) received the bachelor's degree in mathematics and the Ph.D. degree in computer science and engineering from Shanghai Jiao Tong University, China, in 2004 and 2010, respectively. He was a Post-Doctoral Researcher with the National University of Singapore, Singapore, from 2011 to 2012, the University of Illinois at Urbana-Champaign, Champaign, IL, USA, from 2012 to 2013, Cornell University, Ithaca, NY, USA, from 2013 to 2014, and Rutgers University, Piscataway, NJ, USA, in 2014. Since 2014, he has been a Professor with the School of Information and Control, Nanjing University of Information Science and Technology, Nanjing, China. His research interests are pattern recognition and signal processing. He obtained the National Excellent Youth Fund in 2016 and was designated as the global Highly Cited Researchers in 2017.

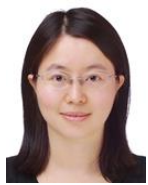

**Dan Zeng** received her Ph.D. degree in circuits and systems in 2008, and her B.S. degree in electronic science and technology in 2003, both from University of Science and Technology of China, Hefei. She is currently a full professor at the Key Laboratory of Specialty Fiber Optics and Optical Access Networks and Shanghai Institute of Advanced Communication and Data Science, Shanghai University, Shanghai. Her research interests include computer vision, multimedia content analysis, and machine learning.

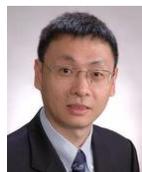

**Shuicheng Yan** (F'16- ) received the Ph.D. degree from the School of Mathematical Sciences, Peking University, in 2004. He is currently the Dean's Chair Associate Professor at National University of Singapore, and also the chief scientist of Qihoo/360 company. Dr. Yan's research areas include machine learning, computer vision and multimedia, and he has authored/ co-authored hundreds of technical papers over a wide range of research topics, with Google Scholar citation over 20,000 times and H-index 66. He is ISI Highly-cited Researcher of 2014-2016. He is an associate editor of IEEE Trans. Knowledge and Data Engineering, IEEE Trans. on Circuits and Systems for Video Technology (IEEE TCSVT) and ACM Trans. Intelligent Systems and Technology (ACM TIST). He received the Best Paper Awards from ACM MM'12 (demo), ACM MM'10, ICME'10 and ICIMCS'09, the winner prizes of classification task in PASCAL VOC 2010-2012, the winner prize of the segmentation task in PASCAL VOC 2012, 2010 TCSVT Best Associate Editor (BAE) Award, 2010 Young Faculty Research Award, 2011 Singapore Young Scientist Award and 2012 NUS Young Researcher Award. He is a Fellow of the IEEE and IAPR.

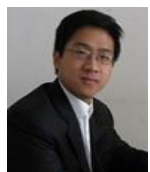

**Meng Wang** is a Professor in the Hefei University of Technology, China. He received the B.E. degree and Ph.D. degree in the Special Class for the Gifted Young and signal and information processing from the University of Science and Technology of China (USTC), Hefei, China, respectively. His current research interests include multimedia content analysis, search, mining, recommendation, and large-scale computing. He has authored 6 book chapters and over 100 journal and conference papers in these areas, including IEEE TMM, TNNLS, TCSVT, TIP, TOMCCAP, ACM MM, WWW, SIGIR, ICDM, etc. He received the paper awards from ACM MM 2009 (Best Paper Award), ACM MM 2010 (Best Paper Award), MMM 2010 (Best Paper Award), ICIMCS 2012 (Best Paper Award), ACM MM 2012 (Best Demo Award), ICDM 2014 (Best Student Paper Award), PCM 2015 (Best Paper Award), SIGIR 2015 (Best Paper Honorable Mention), IEEE TMM 2015 (Best Paper Honorable Mention), and IEEE TMM 2016 (Best Paper Honorable Mention). He is the recipient of ACM SIGMM Rising Star Award 2014. He is/has been an Associate Editor of IEEE Transactions on Knowledge and Data Engineering (TKDE), IEEE Transactions on Neural Networks and Learning Systems (TNNLS) and IEEE Transactions on Circuits and Systems for Video Technology (TCSVT). He is a senior member of the IEEE.